\documentclass[letterpaper]{IEEEtran}
\usepackage{setspace}
\usepackage{graphicx}
\usepackage{amstext}
\usepackage{algorithm}
\usepackage{algorithmic}
\usepackage{mathrsfs}
\usepackage{amssymb}
\usepackage{amsmath}
\usepackage{amsthm}
\usepackage{bm}
\allowdisplaybreaks[4]
\usepackage{epstopdf}
\usepackage{multicol}
\usepackage{stfloats}
\usepackage{url}
\usepackage{color}
\usepackage{enumerate}
\usepackage{comment}
\usepackage{breqn}
\usepackage{bbm}
\usepackage{cite}
\usepackage{subfig}
\usepackage{caption}
\usepackage{enumitem}
\usepackage{multirow}

\DeclareMathAlphabet{\mathscrbf}{OMS}{mdugm}{b}{n}

\usepackage[T1]{fontenc}

\DeclareFontFamily{OML}{zavm}{\skewchar\font=127 }
\DeclareFontShape{OML}{zavm}{m}{it}{<-> s*[.78] zavmri7m}{}
\DeclareFontShape{OML}{zavm}{b}{it}{<-> s*[.78] zavmbi7m}{}
\DeclareFontShape{OML}{zavm}{m}{sl}{<->ssub * zavm/m/it}{}
\DeclareFontShape{OML}{zavm}{bx}{it}{<->ssub * zavm/b/it}{}
\DeclareFontShape{OML}{zavm}{b}{sl}{<->ssub * zavm/b/it}{}
\DeclareFontShape{OML}{zavm}{bx}{sl}{<->ssub * zavm/b/sl}{}
\DeclareMathAlphabet{\mathsf}{OML}{zavm}{m}{it} 

\newtheorem{Lemma}{Lemma}

\newtheorem{Problem}{Problem}
\newtheorem{Definition}{Definition}
\begin{document}
    \title{\huge Joint Optimization of Resource Allocation and Data Selection for Fast and Cost-Efficient Federated Edge Learning}
\author{\IEEEauthorblockN{Yunjian Jia, Zhen Huang, 
 Jiping Yan, Yulu Zhang, Kun Luo, and Wanli Wen} 

\thanks{
The work is sponsored by the National Natural Science Foundation of China under Grant 62201101, the Project funded by China Postdoctoral Science Foundation under Grant 2022M720020, the Natural Science Foundation of Chongqing, China  under Grant cstc2021jcyj-msxmX0458, and the Special Key Projects for Technological Innovation and Application Development in Chongqing Municipality under Grant CSTB2022TIAD-KPX0059.

The authors are with the School of Microelectronics and Communication Engineering, Chongqing University, Chongqing 400044, China (yunjian@cqu.edu.cn, ZhenHuang@cqu.edu.cn, jiping_yan@stu.cqu.edu.cn, yulu_zhang@stu.cqu.edu.cn, kunluo@cqu.edu.cn, wanli_wen@cqu.edu.cn). \emph{(Corresponding author: Wanli Wen)}



%
}

}

\maketitle
\begin{abstract}
Deploying federated learning at the wireless edge introduces federated edge learning (FEEL). Given FEEL’s limited communication resources and potential mislabeled data on devices, improper resource allocation or data selection can hurt convergence speed and increase training costs. Thus, to realize an efficient FEEL system, this paper emphasizes jointly optimizing resource allocation and data selection. Specifically, in this work, through rigorously modeling the training process and deriving an upper bound on FEEL’s one-round convergence rate, we establish a problem of joint resource allocation and data selection, which, unfortunately, cannot be solved directly. Toward this end, we equivalently transform the original problem into a solvable form via a variable substitution and then break it into two subproblems, that is, the resource allocation problem and the data selection problem. The two subproblems are mixed-integer non-convex and integer non-convex problems, respectively, and achieving their optimal solutions is a challenging task. Based on the matching theory and applying the convex-concave procedure and gradient projection methods, we devise a low-complexity suboptimal algorithm for the two subproblems, respectively. Finally, the superiority of our proposed scheme of joint resource allocation and data selection is validated by numerical results.


\end{abstract}

\setlength{\textfloatsep}{5pt} 

\section{Introduction}

It is estimated that there will be 29.3 billion networked devices, such as smart phones, pads, wearable devices and other consumer electronics, by 2024 around the world \cite{cisco2020global}. These devices will inevitably generate huge amounts of data at the wireless edge that can be applicable for multifarious machine learning (ML) tasks, e.g., autonomous driving and product recommendation. Traditional ML algorithms need to expose the raw data to third-party entities for model training, which, however, may compromise data privacy \cite{FLsurvey2020}. To tackle such privacy issue, researchers in the field of wireless communications integrate federated leaning with mobile edge computing, thus forming a concept known as federated edge learning (FEEL)~\cite{review2021Lo,rewardCheng2022,health,vehicle}.
In the training process of FEEL, the devices must send their local training results such as gradients or model parameters, instead of the raw data, via wireless channels. Since the available radio resources such as bandwidth and time at the network edge are constrained, it is necessary to allocate appropriate radio resources for each device during model training. In addition, the data owned by the devices may be mislabeled in practice, for example, a hand-written digit ``1'' may be labeled as ``0'', and an image ``t-shirt'' may be labeled as ``trouser''. Training ML models on such mislabeled data can seriously deteriorate the convergence of FEEL, so it is necessary to select local data appropriately during model training\cite{mcmahan2017communication}.

To date, there have been many research efforts on the resource allocation for FEEL, among which some representative researches are \cite{resall2020huang, luo2021cost, resall2021Mo,NomaFLMa2020, joint2020yu, joint2020Shi, joint2021Wadu,cao2021optimized, joint2021Chen, resall2021Yang, resall2021Dinh, resall2022Wen, NomaFLWu2022, cost2021Lin, cost2019Feng, reward2022Lim}.
Specifically, the authors in~\cite{resall2020huang,luo2021cost,resall2021Mo, NomaFLMa2020, joint2020yu, joint2020Shi, joint2021Wadu,cao2021optimized} developed an efficient joint device scheduling and wireless resource allocation scheme respectively, aiming to minimize devices' total energy cost \cite{resall2020huang, resall2021Mo, luo2021cost} and learning time cost \cite{luo2021cost}, maximize the weighted sum data rate \cite{NomaFLMa2020}, or speed up the convergence of FEEL \cite{joint2020yu, joint2020Shi,joint2021Wadu, cao2021optimized, joint2021Chen}.
The authors in \cite{resall2021Yang} and \cite{resall2021Dinh} established a resource allocation problem to reduce the weighted sum of FEEL training time and total energy cost of all devices.
A joint device scheduling and resource management strategy is developed in \cite{resall2022Wen}, which can significantly speed up model training and save energy costs. Additionally, a joint optimization of the processing-rate, the uplink Non-orthogonal Multiple Access (NOMA) transmission duration, and the broadcasting duration, as well as the accuracy of the local training was proposed in \cite{NomaFLWu2022}, with the aim of minimizing the system-wise cost including the total energy consumption and the FEEL convergence latency.
It is worth noting that excessive energy cost may prevent devices from participating in model training, thus reducing the performance of FEEL. However, how to deal with this issue has not been well studied in the literature \cite{resall2020huang, luo2021cost, resall2021Mo,NomaFLMa2020, joint2020yu, joint2020Shi, joint2021Wadu,cao2021optimized, joint2021Chen, resall2021Yang, resall2021Dinh, resall2022Wen, NomaFLWu2022}. There are several ways to encourage devices to join in the model training process, such as rewarding the devices appropriately to compensate for their energy costs, as done in \cite{cost2021Lin, cost2019Feng, reward2022Lim}.
More specifically, the authors in \cite{cost2021Lin, cost2019Feng, reward2022Lim} proposed to reward all devices based on the number of CPU cycles \cite{cost2021Lin} or the quantity of available training samples \cite{cost2019Feng, reward2022Lim} that they are willing to contribute, and the greater the contribution, the higher the reward. On this basis, they further focused on how to achieve the desirable resource allocation.  Note that the above works have never considered the design of data selection in the FEEL system, so the training algorithm they proposed may not be applicable for scenarios in which some local data are mislabeled.

As for the design of the data selection in FEEL, there are not many works in this direction recently, and some representative works include \cite{sample2021Li, noisy2021Tuor, data-SGD2020He, dataimp2021Aiman, finedata2021Aiman}. Specifically, the work in \cite{sample2021Li} demonstrated the negative influence of data mislabeling on training performance and proposed an efficient data selection method. The authors in \cite{noisy2021Tuor} first evaluated the relevance of data samples and then proposed to filter out all irrelevant data before model training.
In \cite{data-SGD2020He, dataimp2021Aiman, finedata2021Aiman}, the authors built a joint data selection and resource allocation problem to either speed up model training \cite{data-SGD2020He} or minimize the energy cost of FEEL \cite{dataimp2021Aiman, finedata2021Aiman}.
Note that although the above works in \cite{sample2021Li, noisy2021Tuor, data-SGD2020He, dataimp2021Aiman, finedata2021Aiman} have proposed several efficient schemes to speed up model training and reduce the cost of the FEEL system from various perspectives, there are still some limitations. \emph{First}, the works in \cite{sample2021Li} and \cite{noisy2021Tuor} overlooked the resource allocation of the FEEL system. \emph{Second}, the authors in~\cite{data-SGD2020He, dataimp2021Aiman, finedata2021Aiman} did not analyze the convergence of the training process. \emph{Third}, the studies in \cite{sample2021Li, noisy2021Tuor, data-SGD2020He, dataimp2021Aiman, finedata2021Aiman} assumed that all devices can always participate in model training, which, however, may not hold in practice, since the edge devices may fail to upload local gradients due to the loss of connections. As a result, the proposed schemes therein may not be suitable for practical FEEL systems.

In this work, we would like to address the above issues.
We investigate a generic FEEL system consisting of multiple devices and an edge server, where each device connect to the server via wireless channels. It is worth noting that to simulate the practical scenarios, we assume that the devices may not always be available to connect to the server.
In addition, some devices may be not willing to conduct model training due to the high energy cost. Thus, to entice devices to join in model training, the server will reward the devices to compensate for the energy cost. Particularly,  in the FEEL system, we consider that the communication resources are limited and each device may have some mislabeled data samples.
On these bases, we would like to analyze the convergence of FEEL and jointly optimize the resource allocation and data selection to speed up model training while reducing the net cost (i.e., the energy cost minus the reward) of all devices.
Our contributions are listed below.

\begin{itemize}
    \item{By mathematically modeling the training process of FEEL, we derive an analytical expression for the net cost of all devices. In addition, we propose a new gradient aggregation method and derive an upper bound on FEEL's one-round convergence rate. On this basis, we obtain a mathematical expression that can be used to speed up the convergence of FEEL.}
    \item{To accelerate the convergence of FEEL while minimizing the net cost of all devices, we establish a joint resource allocation and data selection problem.  Since the formulated problem is unsolvable on the server side, we equivalently transform it into a more solvable form with some appropriate transformations and then separate the original problem into two subproblems: the \textit{resource allocation problem} and the \textit{data selection problem}. These two subproblems are mixed-integer non-convex and integer non-convex problems, respectively, and it is very challenging to obtain their optimal solutions.}
    \item{Based on the matching theory and applying the convex-concave procedure and gradient projection methods, we propose a low-complexity suboptimal algorithm for the resource allocation problem and the data selection problem, respectively.}
\end{itemize}

\section{System Model}\label{sectionSystemModel}
\begin{figure}[t]
  \centering
  \includegraphics[width= 0.5\textwidth]{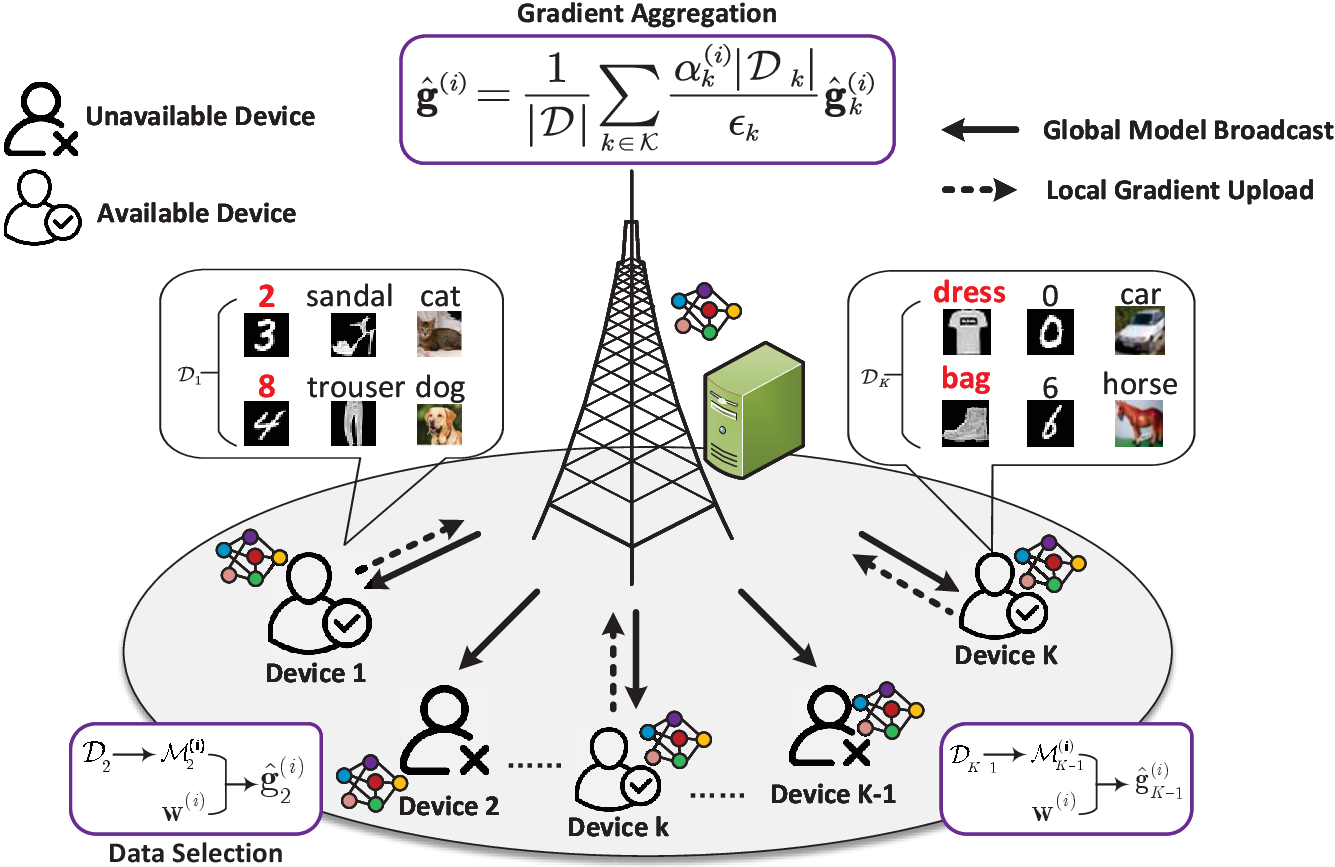}
  \caption{System model. Device $k$ has $|\mathcal{D}_k| = 6$ images, two of which are mislabeled. The text above each image represents the corresponding label, with black text and red text representing correct and incorrect labels, respectively.}\label{figsystemmodel}
  \vspace{3mm}
\end{figure}

We investigate a generic FEEL system, which consists of one  edge server and $K$ devices, as depicted in Fig.~\ref{figsystemmodel}. {To maintain the simplicity of analysis and optimization, we assume that both the BS and devices are equipped with a single antenna.} Let $\mathcal{K}=\left\{1,2,\cdots, K\right\}$ denote the user set. {Note that we assume that all devices are legitimate, with no malicious devices.} Let $\mathcal{D}_k=\{(\mathbf{x}_j,y_j)\}_{j=1}^{|\mathcal{D}_k|}$ denote the dataset of the $k$-th device. Here, $\mathbf{x}_j\in\mathbb{R}^d$ is the $d$-dimensional data sample, $y_j\in\mathbb{R}$ denotes the data label, and $\left|\cdot\right|$ depicts the cardinality of a set. We define $\mathcal{D}= \cup_{k\in \mathcal{K}}\mathcal{D}_k$ as the whole dataset resided on all devices. Table~\ref{table1} summarized the main notations used in this paper.

\begin{table*}[ht]\vspace{-5mm}
\small
\caption{List of Notations}\label{table1}
\centering
\begin{tabular}{l|p{12cm}lp{12cm}}
    \hline
    \textbf{Notation} & \textbf{Definition} \\
    \hline
    $\mathcal{N}$; $\mathcal{K}$; $N$; $K$ & Set of resource blocks (RBs); set of devices; number of RBs; number of devices \\
    \hline
    $\mathcal{D}_k$; $\mathcal{D}$; $\left|\mathcal{D}_k\right|$; $\left|\mathcal{D}\right|$ &  Dataset on device $k$; the whole dataset; cardinality of $\mathcal{D}_k$; cardinality of $\mathcal{D}$ \\
    \hline
    $\mathbf{w}^{(i)}$; $\mathbf{x}_j$; $y_j$ &  ML model at round $i$; data sample $j$; label of data sample $j$ \\
    \hline
    $L(\cdot)$; $L_k(\cdot)$; $\ell(\cdot)$ & Global loss function; local loss function of device $k$; loss function of each data sample\\
    \hline
    $\hat{\mathbf{g}}^{(i)}$; $\hat{\mathbf{g}}_k^{(i)}$; $\mathbf{g}_{k,j}^{(i)}$  & Global gradient at round $i$; local gradient of device $k$ at round $i$; gradient of the $j$-th data sample on device $k$ at round $i$ \\
    \hline
    $q_k$; $c_k$ & Reward per data sample for device $k$; cost per unit energy consumption of device $k$ \\
    \hline
    $F_k$; $f_k$; $\kappa$ & Number of CPU cycles required for computing $\mathbf{g}_{k,j}^{(i)}$; CPU frequency of device $k$; energy capacitance coefficient \\
    \hline
    $R(\cdot)$; $C^{\rm cmp}$; $C^{\rm com}$; $C(\cdot)$ &  Total reward received by all devices;  total cost of performing gradient computing on all devices; total cost of performing local gradient uploading from all devices; net cost of all devices \\
    \hline
    $\alpha_k^{(i)}$; $\epsilon_k$ & Availability indicator of device $k$ at round $i$; probability that device $k$ is available for gradient uploading \\
    \hline
    $B$; $L$; $T$; $r_{k,n}^{(i)}$ & Bandwidth of each RB; size of gradient; time duration for uploading; achievable data rate of device $k$ on RB $n$ at round $i$ \\
    \hline
    $\boldsymbol{\mathcal{M}}^{(i)}$; $\boldsymbol{\rho}^{(i)}$; $\mathbf{p}^{(i)}$ & Data selection at round $i$; RB assignment at round $i$; power allocation at round $i$\\
    \hline
\end{tabular}
\end{table*}

\subsection{Learning Model}  
The FEEL system can learn an ML model $\mathbf{w}\in\mathbb{R}^d$ over the datasets of all devices by solving the problem below
\begin{align}
\mathbf{w}^*= \arg\min_{\mathbf{w}}L(\mathbf{w}).
\end{align}
Here, $L(\mathbf{w})=\frac{1}{\left| \mathcal{D} \right|}\sum_{k\in\mathcal{K}}{\left| \mathcal{D}_k \right|L_k(\mathbf{w})}$ is the loss function with
\begin{align}\label{Lk}
L_k(\mathbf{w}) =\frac{1}{\left| \mathcal{D}_k \right|}\sum_{j\in \mathcal{D}_k}{\ell(\mathbf{w},\mathbf{x}_j,y_j)},
\end{align}denoting the loss function for device $k$ and $\ell(\cdot)$ being an appropriate sample-wise loss function.

Training an ML model in the FEEL system usually is an iterative procedure, in which the iteration is also known as the communication round. Define $\mathbf{w}^{(i)}$ to be the global model at round $i=1,2,\cdots$. Then, the training process of FEEL at each round consists of three stages, i.e., \emph{Local Gradient Computing}, \emph{Local Gradient Uploading}, and \emph{Global Model Updating}. We will detail these stages in the following subsections.

\subsection{Local Gradient Computing}
The $k$-th device will compute the local gradient, denoted by $\hat{\mathbf{g}}_k^{(i)}$, of $L_k(\mathbf{w})$ at $\mathbf{w}=\mathbf{w}^{(i)}$, where $\mathbf{w}^{(i)}$ is received from the edge server. The computed gradient $\hat{\mathbf{g}}_k^{(i)}$ is uploaded to the server subsequently in the stage of \emph{Local Gradient Uploading}. Note that, most of the current literature on FEEL implicitly assumes that the device's data samples are always labeled correctly \cite{resall2020huang, NomaFLMa2020, resall2021Yang, resall2021Mo, resall2021Dinh, joint2020yu, joint2020Shi, joint2021Wadu, joint2021Chen, NomaFLWu2022, resall2022Wen}, and thus the gradient $\hat{\mathbf{g}}_k^{(i)}$ is
\begin{align}\label{gki}
\hat{\mathbf{g}}_{k}^{(i)} = \nabla L_k(\mathbf{w}^{(i)}) = \frac{1}{\left|\mathcal{D}_k\right|}\sum_{(\mathbf{x}_j,y_j) \in \mathcal{D}_{k}}\mathbf{g}_{k,j}^{(i)}, \ {\forall k\in\mathcal{K}}
\end{align}where $\mathbf{g}_{k,j}^{(i)} = \nabla \ell(\mathbf{w}^{(i)},\mathbf{x}_j,y_j)$ and $L_k(\mathbf{w}^{(i)})$ is given by (\ref{Lk}). Nevertheless, in practice, some data samples in $\mathcal{D}_k$ may be mislabeled, for example, a hand-written digit ``1'' may be labeled as ``0'', and an image ``t-shirt'' may be labeled as ``trouser''. Such mislabeling may degenerate the training performance of FEEL if directly sending $\mathbf{\hat{g}}_{k}^{(i)}$ in (\ref{gki}) to the edge server. To this end, we consider a data selection design,\footnote{Note that the data selection mechanism discussed in this paper does not necessitate the capability of the user devices or the edge server to detect data with incorrect labels.} in which only a subset $\mathcal{M}_k^{(i)}$ of the data samples in $\mathcal{D}_k$ are selected to compute the local gradient.
Specifically, a subdataset is first sampled from $\mathcal{D}_k$ with size $|\hat{\mathcal{D}}_k|$, where $\hat{\mathcal{D}}_k \subseteq \mathcal{D}_k$. Then, the gradient norm square of each data sample in $\hat{\mathcal{D}}_k$ is computed and sent to the BS. On this basis, a subset $\mathcal{M}_k^{(i)}$ is chosen from $\hat{\mathcal{D}}_k$ to compute the local gradient.
On this basis, with a slight abuse of notation, we reexpress (\ref{gki}) as
\begin{align}\label{gki2}
\mathbf{\hat{g}}_{k}^{(i)} = \frac{1}{|\mathcal{M}_k^{(i)}|}\sum_{(\mathbf{x}_j,y_j) \in \mathcal{M}_{k}^{(i)}}\mathbf{g}_{k,j}^{(i)}, \ {\forall k\in\mathcal{K}}
\end{align}where $\mathcal{M}_k^{(i)}$ satisfies the following constraints
\begin{align}
& \mathcal{M}_k^{(i)}\subseteq \hat{\mathcal{D}}_k,\quad {\forall}k\in \mathcal{K},\label{consOfDataSel01}\\
& \mathcal{M}_k^{(i)} \neq \emptyset,\quad {\forall}k\in \mathcal{K}.\label{consOfDataSel02}
\end{align}Let $\boldsymbol{\mathcal{M}}^{(i)} = \{ \mathcal{M}_k^{(i)}\}_{k\in\mathcal{K}}$ denote the data selection design.
{Note that the data selection design $\boldsymbol{\mathcal{M}}^{(i)}$ can be applicable in smart healthcare systems for disease diagnosis \cite{health} and in autonomous vehicles for enhanced driving process safety \cite{vehicle}.}
After completing the local gradient calculation, device $k$ will send the local gradient $\mathbf{\hat{g}}_{k}^{(i)} $ given in (\ref{gki2}) instead of (\ref{gki}) to the edge~server.

As done in \cite{cost2019Feng} and \cite{rewardCheng2022}, to compensate for the cost of user participation in training and attract them to participate in model training, we consider that the server will reward each device.
Let $q_k$ denote the reward per data sample for device $k$.\footnote{Determining the specific value of $q_k$ is a strategic process that takes into account a range of factors, including but not limited to data volume and privacy concerns.} Then, the total reward received by all devices in $\mathcal{K}$ is
\begin{align}\label{eqFee}
R(\boldsymbol{\mathcal{M}}^{(i)}) = \sum_{k\in\mathcal{K}} q_k|\mathcal{M}_k^{(i)}|.
\end{align}On the other hand, the local calculation will consume a certain amount of energy of each device.
{To be specific, we define $F_k$ as the CPU computational power required by device $k$ to compute the gradient norm square of a single data sample, which is quantified in terms of CPU cycles per sample. Concurrently, let $f_k$ denote the CPU frequency of device $k$, measured in CPU cycles per second.} Then, the computation time at device $k$ for computing the gradient norm square of the data samples in $\hat{\mathcal{D}}_k$ is given by
\begin{align}
    \tau_k=\frac{F_k |\hat{\mathcal{D}}_k|}{f_k}, \ {\forall}k\in\mathcal{K}.
\end{align}Based on \cite{cmp2016Mao}, the energy consumption for gradient computing at device $k$ is
\begin{align}
    E_k^{\rm{cmp}} = \kappa F_k |\hat{\mathcal{D}}_k|f_k^2, \ {\forall k\in\mathcal{K}}.
\end{align}Here $\kappa$ denotes the energy capacitance coefficient which is dependent on the chip architecture.
Let $c_k$ denote the cost per unit energy consumption of device $k$.\footnote{Note that accurately quantifying $c_k$ in real-world FEEL systems is complex, as it involves various factors including the energy efficiency of the hardware, the energy required for data transfer, and the operational efficiency of the model on specific devices \cite{luo2021cost}.} Thus, the total cost of performing gradient computing on all devices is given by
\begin{align} \label{eqEnrCmp}
C^{\rm cmp}= \sum_{k\in\mathcal{K}}{  c_k E_k^{\rm{cmp}} }.
\end{align}

\subsection{Local Gradient Uploading}
In this stage, each device needs to submit its gradient to the edge server. Due to the resource limitation of wireless transmission, we consider a {grant-based} NOMA system that has a set $\mathcal{N}=\left\{1,2,\cdots,N\right\}$ of $N$ resource blocks (RBs). To support the gradient uploading, these RBs in $\mathcal{N}$ should be carefully allocated to the devices in $\mathcal{K}$. Define $\rho_{k,n}$ to be the RB assignment indicator, where $\rho_{k,n}=1$ indicates that the $n$-th RB is assigned to the $k$-th device, and $\rho_{k,n}=0$, otherwise. Consider that each RB can be allocated to $Q\ge1$ devices at most, while each device can occupy only one RB \cite{noma2019Han, NomaOma2019}, i.e., the following constraints should be satisfied
\begin{align}
&\rho_{k,n}^{(i)}\in \{0,1\},~{\forall}k \in \mathcal{K},~{\forall}n \in \mathcal{N},\label{consOfrho01}\\
&\sum_{k\in \mathcal{K}}{\rho_{k,n}^{(i)}} \le Q, ~{\forall}n\in \mathcal{N},\label{consOfrho02}\\
&\sum_{n\in\mathcal{N}}{\rho_{k,n}^{(i)}} \le 1,~{\forall}k\in \mathcal{K}.\label{consOfrho03}
\end{align}Additionally, as mentioned earlier, not all devices are available to transmit their gradients to the server. To reflect such behavior, let $\alpha_k \in \{0,1\}$ denote device $k$'s availability state, where $\alpha_k = 1$ indicates that device $k$ can be available to upload its gradient $\hat{\mathbf{g}}_k^{(i)}$, and $\alpha_k = 0$, otherwise. Then, we reform the uploaded local gradient of device $k$ as $\alpha_k^{(i)}\hat{\mathbf{g}}_k^{(i)}$, and the following constraint should be satisfied
\begin{align}
\rho_{k,n}^{(i)} \le \alpha _k^{(i)},~{\forall}k\in \mathcal{K},\; {\forall}n\in\mathcal{N}.\label{consOfrho04}
\end{align}

According to the NOMA transmission protocol \cite{zhu2022efficient,zhang2023reinforcement,shi2021multi,shi2022sparse}, the uplink signals from the devices occupying the same RB will be superposed. Let $\mathcal{S}_n^{(i)}$ denote the set of devices occupying RB $n$. Then, the received signal at the edge server on RB $n$ can be expressed as
\begin{align}
u_n^{(i)}=\sum_{k\in \mathcal{S}_n^{(i)}}{\sqrt{\rho_{k,n}^{(i)}p_{k,n}^{(i)}h_{k,n}^{(i)}}\alpha_k^{(i)} v_{k,n}^{(i)}}+ m, \ {\forall n\in\mathcal{N}}\nonumber
\end{align}
where $v_{k,n}^{(i)}$ denotes the signal transmitted from device $k$ on the $n$-th RB with $\mathbb{E}\{| v_{k,n}^{(i)}|^2 \} =1$ and $\mathbb{E}\{\cdot\}$ being the expectation operator, $h_{k,n}^{(i)}$ is the channel power gain of device $k$ on RB $n$, $m$ denotes the zero-mean Gaussian noise with variance $N_0$, $p_{k,n}^{(i)}$ denotes the transmission power of device $k$ on RB $n$. Let $p_k^{\rm{max}}$ represent the maximum power limit of device $k$. Then, we have
\begin{align}
0\le p_{k,n}^{(i)}\le \rho _{k,n}^{(i)}p_{k}^{\rm max},\ {\forall}k\in \mathcal{K},\ {\forall}n\in \mathcal{N}\label{consOfMaxPow01},
\end{align}
We consider that the successive interference cancellation is applied at the edge server. Specifically, for each RB, the edge server first decodes the signal from the device with the highest channel power gain and treats the signals of others as interference. Then, the edge server subtracts the decoded signal from the superposed signal. {This iterative process continues until the signals from all users have been successfully decoded.}
On this basis, we can calculate the achievable rate of device $k$ on RB $n$ as follows, $r_{k,n}^{(i)}=B\log_2 \left(1+\frac{\rho_{k,n}^{(i)} p_{k,n}^{(i)} h_{k,n}^{(i)}}{\sum_{t\in \mathcal{S}_n^{(i)}}{\mathbbm{1}\left[ h_{t,n}^{(i)}<h_{k,n}^{(i)} \right]\rho_{t,n}^{(i)}p_{t,n}^{(i)}h_{t,n}^{(i)}} + N_0}\right), \quad {\forall}k \in \mathcal{K}, \quad {\forall}n \in \mathcal{N}$.
 Here, $B$ represents the bandwidth of each RB and $\mathbbm{1}\left[ \cdot \right]$ denotes an indicator function. Let $L$ and $T$ represent the size of the local gradient $\mathbf{\hat{g}}_{k}^{(i)} $ in (\ref{gki2}) and the time duration for gradient uploading, respectively. To ensure that $\mathbf{\hat{g}}_{k}^{(i)}$ can be uploaded to the server successfully, we have
\begin{align}
\sum_{n\in\mathcal{N}}{r_{k,n}^{(i)}}T\ge \alpha_k^{(i)}L,~{\forall}k\in \mathcal{K}.\label{consOfRate01}
\end{align}

Finally, the energy consumption of device $k$ for local gradient uploading is $E^{\rm{com}}_k\left(\boldsymbol{\rho}^{(i)},\mathbf{p}^{(i)}\right)=\sum_{n\in\mathcal{N}}{\rho_{k,n}^{(i)}}p_{k,n}^{(i)}T$, and the total cost of performing local gradient uploading from all devices can be expressed as
\begin{align}\label{eqEnrCom}
C^{\rm com}\left(\boldsymbol{\rho}^{(i)},\mathbf{p}^{(i)}\right) = \sum_{k\in\mathcal{K}}{c_kE_k^{\rm{com}}\left(\boldsymbol{\rho}^{(i)},\mathbf{p}^{(i)}\right)},
\end{align}
where $\boldsymbol{\rho}^{(i)}=(\rho_{k,n}^{(i)})_{k\in\mathcal{K}, n\in \mathcal{N}}$ and $\mathbf{p}^{(i)}=(p_{k,n}^{(i)})_{k\in\mathcal{K}, n\in\mathcal{N}}$ denote the design parameters of RB assignment and transmission power allocation, respectively. Finally, based on the reward in (\ref{eqFee}), and the total costs in (\ref{eqEnrCmp}) and (\ref{eqEnrCom}),  we can compute the net cost (i.e., the energy cost minus the reward) of all devices as follows.
\begin{align}\label{profitOfDev}
C(\boldsymbol{\mathcal{M}}^{(i)}, \boldsymbol{\rho}^{(i)},\mathbf{p}^{(i)}) = C^{\rm com}\left(\boldsymbol{\rho}^{(i)},\mathbf{p}^{(i)}\right)+ C^{\rm cmp} - R(\boldsymbol{\mathcal{M}}^{(i)}).
\end{align}

\subsection{Global Model Updating}
In this stage, by aggregating the gradients from the devices, the edge server will generate a new global model for the next round of model training.
Let $\epsilon_k = \Pr[\alpha_k^{(i)} = 1]$ denote the probability that device  $k$ is available to submit its local gradient. Then, we propose that at round $i$, the server aggregates the local gradients based on
\begin{align}
\mathbf{\hat{g}}^{(i)}=\frac{1}{|\hat{\mathcal{D}}|}\sum_{k \in \mathcal{K}}{\frac{|\hat{\mathcal{D}}_k|}{\epsilon _k}\alpha _{k}^{(i)}\mathbf{\hat{g}}_{k}^{(i)}},\label{eqagggra}
\end{align}where $\hat{\mathcal{D}}=\sum{_{k=1}^K \hat{\mathcal{D}}_k}$, and $\hat{\mathbf{g}}^{(i)}$ denotes the global gradient generated on the edge server side.

The following lemma shows the unbiasedness of $\mathbf{\hat{g}}^{(i)}$, which can greatly facilitate us to prove FEEL's convergence rate.
\begin{Lemma}\label{lemmaOfUnbias}
The expectation of $\hat{\mathbf{g}}^{(i)}$ is equal to the ground-truth gradient $\mathbf{g}^{(i)} = \nabla L(\mathbf{w}^{(i)})$.
\end{Lemma}

The proof can be found in Appendix~\ref{proofOfLemma1}. Based on (\ref{eqagggra}), the edge server generates a new global model $\mathbf{w}^{(i+1)}$ for the next round of model training according to
\begin{align}
\mathbf{w}^{(i+1)}=\mathbf{w}^{(i)}-\eta^{(i)} \mathbf{\hat{g}}^{(i)}.
\end{align}
Here, $\eta^{(i)}>0$ denotes the learning rate at round $i$.

\subsection{Convergence Analysis}
The above three stages will be repeated several times until FEEL converges. We now analysis the convergence behaviour of FEEL under the proposed data selection scheme.
\subsubsection{One-round convergence rate}
Using Lemma~\ref{lemmaOfUnbias}, we first derive an upper bound on FEEL's one-round convergence rate.
\begin{Lemma}\label{lemmaOfConvRate}
If $\nabla L(\mathbf{w})$ is Lipschitz continuous, we obtain
\begin{align}\label{convBound}
&\mathbb{E}\left[ L\left(\mathbf{w}^{(i+1)}\right) -L( \mathbf{w}^*)\right] \le \mathbb{E}\left[ L\left(\mathbf{w}^{(i)}\right) -L(\mathbf{w}^*)\right]\nonumber \\
    &-\eta^{(i)} \left\| \mathbf{g}^{(i)} \right\| ^2
    +\frac{\beta \left(\eta^{(i)}\right) ^2}{2|\hat{\mathcal{D}}|^2}\Delta(\boldsymbol{\mathcal{M}}^{(i)}).
\end{align}
Here, $\beta>0$ is the Lipschitz modulus and $\Delta(\boldsymbol{\mathcal{M}}^{(i)})$ is given by (\ref{expOfDelta}), shown at the bottom of next page, where $\sigma_{k,j}^{(i)}=\| \mathbf{g}_{k,j}^{(i)} \|^2$.

\begin{figure*}[hb]\vspace{-5mm}
\hrulefill
\begin{footnotesize}
\begin{align}\label{expOfDelta}
\Delta(\boldsymbol{\mathcal{M}}^{(i)}) =  \sum_{k\in \mathcal{K}}{\left(\frac{|\hat{\mathcal{D}}_k|^2}{\epsilon_k |\mathcal{M}_k^{(i)}|}\sum_{(\mathbf{x}_j,y_j)\in \mathcal{M}_k^{(i)}}{\sigma_{k,j}^{(i)}}+\sum_{t\in \mathcal{K} \setminus \{ k\}}{\frac{|\hat{\mathcal{D}}_k||\hat{\mathcal{D}}_t|}{|\mathcal{M}_t^{(i)}|}\sum_{(\mathbf{x}_j,y_j)\in \mathcal{M}_t^{(i)}}{\sigma _{t,j}^{(i)}}}\right)}.
\end{align}
\end{footnotesize}
\end{figure*}

\end{Lemma}

The proof can be found in Appendix~\ref{proofOfLemma2}.

\subsubsection{Convergence upper bound}
Based on Lemma \ref{lemmaOfConvRate}, we can derive the convergence upper bound in the $i$-th communication round.

\begin{Lemma}\label{lemmaOfConvBnd}
If $\nabla L(\mathbf{w})$ is strongly convex with a positive parameter $\mu$, we have
\begin{align}\label{ConvBnd}
    &\mathbb{E}[L(\mathbf{w}^{(i+1)})-L(\mathbf{w}^{*})]
    \le \prod \limits_{t=1}^i (1-2\mu\eta^{(t)})
    \mathbb{E}[L(\mathbf{w}^{(1)})-L(\mathbf{w}^*)]\nonumber \\
    &+\frac{\beta}{2|\hat{\mathcal{D}}|^2}\sum_{t=1}^i{A^{(t)}(\eta^{(t)})^2\Delta(\boldsymbol{\mathcal{M}}^{(t)})},
\end{align}where $A^{(t)}=\prod \limits_{j=t+1}^i (1-2\mu \eta^{(j)})$ is a weight coefficient.
\end{Lemma}

The proof can be found in Appendix~\ref{proofOfLemma3}. Lemmas \ref{lemmaOfConvRate} and \ref{lemmaOfConvBnd} indicate that $\Delta(\boldsymbol{\mathcal{M}}^{(i)})$ directly relates to the data selection $\boldsymbol{\mathcal{M}}^{(i)}$ and the decrease of $\Delta(\boldsymbol{\mathcal{M}}^{(i)})$ will speed up the convergence of FEEL.


\section{Problem Formulation and Transformation}\label{sectionProbFormu}

In the sequel, we first establish a joint resource allocation and data selection problem. Then, we conduct some necessary transformations to facilitate solving the original problem.

\subsection{Problem Formulation}
According to (\ref{profitOfDev}) and (\ref{convBound}), we see that the variables $(\boldsymbol{\rho}^{(i)},\mathbf{p}^{(i)})$ and $\boldsymbol{\mathcal{M}}^{(i)}$ can affect both the convergence of FEEL and the net cost of all devices. Therefore, a question naturally arises: how to design an appropriate joint resource allocation and data selection scheme to accelerate the convergence of FEEL while minimizing the net cost of all devices?
To answer this question, the following problem is established.

\begin{Problem}[Joint Resource Allocation and Data Selection]\label{OriginalProblem}
\begin{align}
\min_{\boldsymbol{\mathcal{M}}^{(i)},\boldsymbol{\rho}^{(i)},\mathbf{p}^{(i)}}\ & \lambda \Delta(\boldsymbol{\mathcal{M}}^{(i)}) +(1-\lambda) C(\boldsymbol{\mathcal{M}}^{(i)}, \boldsymbol{\rho}^{(i)},\mathbf{p}^{(i)}) \nonumber \\
\mathrm{s.t.}\quad & (\ref{consOfDataSel01}), (\ref{consOfDataSel02}), (\ref{consOfrho01}), (\ref{consOfrho02}), (\ref{consOfrho03}), (\ref{consOfrho04}), (\ref{consOfMaxPow01}), (\ref{consOfRate01}).\nonumber
\end{align}
\end{Problem}


Problem~\ref{OriginalProblem} should be tackled on the server side, not on the device side. In particular, due to the constraints in (\ref{consOfDataSel01}) and (\ref{consOfDataSel02}), all users' local data should be uploaded to the server when solving Problem~\ref{OriginalProblem}. However, to protect data privacy, FEEL does not allow the server to directly access the raw data of each device, so Problem~\ref{OriginalProblem} cannot be solved at the edge server. {In the following subsection, we will transform Problem~\ref{OriginalProblem} into a solvable form through variable substitution. Subsequently, we will decompose it into two subproblems. Due to the transformation, the server can address Problem~\ref{OriginalProblem} without compromising user data confidentiality.}

\subsection{Problem Transformation}
Let $\delta_{k,j}^{(i)}\in \{0,1\}$ denote the data selection indicator, where $\delta_{k,j}^{(i)} = 1$ represents that the $j$-th data sample in $\mathcal{D}_k$ is selected into the set $\mathcal{M}_k^{(i)}$, and $\delta_{k,j}^{(i)} = 0$, otherwise.
Define the variable $\boldsymbol{\delta}^{(i)}=(\boldsymbol{\delta}_k^{(i)})_{k\in\mathcal{K}}$, where $\boldsymbol{\delta}_k^{(i)}=(\delta_{k,j}^{(i)})_{j\in\mathcal{J}_k}$ and $\mathcal{J}_k = \{1,2,\cdots,|\hat{\mathcal{D}}_k|\}$. Then, by replacing $\mathcal{M}_k$ with $\boldsymbol{\delta}^{(i)}$, we transform Problem~\ref{OriginalProblem} into an equivalent problem as follows.

\begin{Problem}[Joint Resource Allocation and Data Selection]\label{EqualProblem}
\begin{align}
\min_{\boldsymbol{\delta}^{(i)},\boldsymbol{\rho}^{(i)},\mathbf{p}^{(i)}} & \lambda \hat{\Delta}(\boldsymbol{\delta}^{(i)}) + (1-\lambda) \hat{C}(\boldsymbol{\delta}^{(i)},\boldsymbol{\rho}^{(i)},\mathbf{p}^{(i)}) \nonumber \\
\mathrm{s.t.}\ & {(\ref{consOfrho01}), (\ref{consOfrho02}), (\ref{consOfrho03}), (\ref{consOfrho04}), (\ref{consOfMaxPow01}), (\ref{consOfRate01}),} \nonumber \\
& \delta_{k,j}^{(i)}\in \{ 0,1 \},\;\; j\in\mathcal{J}_k,~{\forall}k\in \mathcal{K},\label{consOfDataSelDelta01}\\
& 0 < \sum_{j\in\mathcal{J}_k}{\delta_{k,j}^{(i)}}\le |\hat{\mathcal{D}}_k|,~{\forall}k\in \mathcal{K} \label{consOfDataSelDelta02},
\end{align}
where $\hat{\Delta(\boldsymbol{\delta}^{(i)})}$ is given in (\ref{delta01}), shown at the bottom of next page, and the net cost of devices is given by

\begin{figure*}[hb]\vspace{-5mm}
\hrulefill
\begin{footnotesize}
\begin{align}\label{delta01}
\hat{\Delta}(\boldsymbol{\delta}^{(i)}) = \sum_{k\in\mathcal{K}}{\left(\frac{|\hat{\mathcal{D}}_k|^2}{\epsilon_k\sum_{j\in\mathcal{J}_k}{\delta _{k,j}^{(i)}}}\sum_{j\in\mathcal{J}_k}{\delta_{k,j}^{(i)}\sigma_{k,j}^{(i)}}+\sum_{t\in \mathcal{K}\setminus \{k\}}{\frac{|\hat{\mathcal{D}}_k||\hat{\mathcal{D}}_t|}{\sum_{j\in\mathcal{J}_t}{\delta _{t,j}^{(i)}}}\sum_{j\in\mathcal{J}_t}{\delta _{t,j}^{(i)}\sigma _{t,j}^{(i)}}}\right)}.
\end{align}
\end{footnotesize}

\end{figure*}
\begin{footnotesize}
\begin{align}
\hat{C}(\boldsymbol{\delta}^{(i)},\boldsymbol{\rho}^{(i)},\mathbf{p}^{(i)})=C^{\rm{com}}\left( \boldsymbol{\rho}^{(i)},\mathbf{p}^{(i)} \right)+C^{\rm{cmp}}-\sum_{k\in\mathcal{K}}{q_k\sum_{j\in\mathcal{J}_k}{\delta_{k,j}^{(i)}}}.
\end{align}
\end{footnotesize}
\end{Problem}

Compared to the original Problem~\ref{OriginalProblem}, Problem~\ref{EqualProblem} is a solvable problem, since it is only required to send the cardinality of $\hat{\mathcal{D}}_k$ to the server, instead of $\hat{\mathcal{D}}_k$ itself. Nonetheless, it is still difficult to achieve an optimal point of Problem~\ref{EqualProblem} for the following reasons. First, the variables $\boldsymbol{\rho}^{(i)}$ and $\boldsymbol{\delta}^{(i)}$ are binary. Second, the objective function and the constraint in (\ref{consOfRate01}) are non-convex. Thus, Problem~\ref{EqualProblem} is recognized as a highly challenging mixed-integer non-convex problem. {To tackle Problems~\ref{ResAllProblem}, we decompose Problem~\ref{EqualProblem} into it and \ref{DataSelProblem} equivalently. }

\begin{Problem}[Resource Allocation Problem]\label{ResAllProblem}
\begin{align}
&\min_{\boldsymbol{\rho}^{(i)},\mathbf{p}^{(i)}} C^{\rm com}\left( \boldsymbol{\rho}^{(i)},\mathbf{p}^{(i)} \right)+C^{\rm cmp} \nonumber \\
&\mathrm{s.t.}\quad (\ref{consOfrho01}), (\ref{consOfrho02}), (\ref{consOfrho03}), (\ref{consOfrho04}), (\ref{consOfMaxPow01}), (\ref{consOfRate01}).\nonumber
\end{align}
\end{Problem}

\begin{Problem}[Data Selection Problem]\label{DataSelProblem}
\begin{align}
&\min_{\boldsymbol{\delta}^{(i)}}\ \lambda \hat{\Delta}(\boldsymbol{\delta}^{(i)}) + (1-\lambda) \hat{C}(\boldsymbol{\delta}^{(i)},\boldsymbol{\rho}^{(i)},\mathbf{p}^{(i)}) \nonumber \\
&\mathrm{s.t.}\quad  (\ref{consOfDataSelDelta01}), (\ref{consOfDataSelDelta02}).\nonumber
\end{align}
\end{Problem}



\begin{figure}[ht]
  \centering
  \includegraphics[width= 0.5\textwidth]{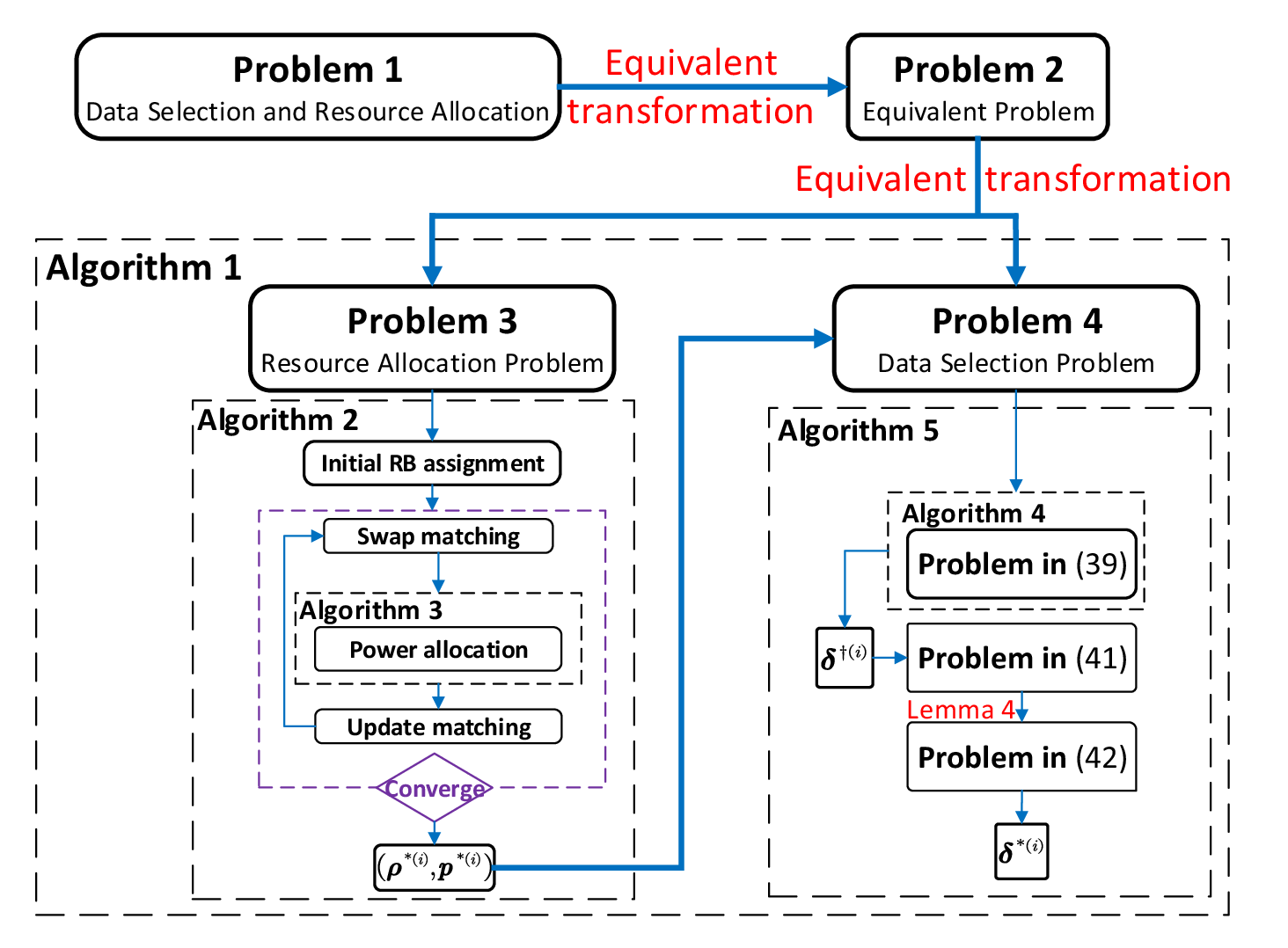}
  \caption{A schematic diagram for solving Problem~\ref{EqualProblem}.}\label{figflowchart}
  \vspace{-3mm}
\end{figure}

In the sequel, we will solve Problem~\ref{EqualProblem} by solving Problem~\ref{ResAllProblem} and Problem~\ref{DataSelProblem}, as illustrated in Fig.~\ref{figflowchart}. The details of solving Problem~\ref{EqualProblem} are summarized in Algorithm~\ref{overallAlgorithm}.\footnote{Note that in Algorithm~\ref{overallAlgorithm}, each user performs a single iteration over their local data to compute the local gradient. Consequently, Algorithm~\ref{overallAlgorithm} can be considered a variant of the widely-employed FedSGD algorithm \cite{mcmahan2017communication} in federated learning. The analysis and optimization presented herein can be adapted to scenarios utilizing the FedAvg algorithm \cite{mcmahan2017communication}, wherein users typically execute multiple iterations on their local data and upload model parameters instead of gradients to the edge server.}
Specifically, the edge server first solves Problem~\ref{ResAllProblem} to obtain the resource allocation design, denoted by $(\boldsymbol{\rho}^{*(i)}, \mathbf{p}^{*(i)})$, via Algorithm~\ref{algresall} (see Section~\ref{sectionSolutionToProblem3}), as shown in Step 2. Then, Problem~\ref{DataSelProblem} is solved to obtain the data selection, denoted by $\boldsymbol{\delta}^{*(i)}$, via Algorithm~\ref{algdatsel} (see Section~\ref{sectionSolutionToProblem4}), as shown in Step 3. Note that, the edge server sends $(\boldsymbol{\delta}^{*(i)}, \boldsymbol{\rho}^{*(i)}, \mathbf{p}^{*(i)})$ back to all devices in $\mathcal{K}$, and each device can then determine $\boldsymbol{\mathcal{M}}^{*(i)}$ according to $\boldsymbol{\delta}^{*(i)}$ and select all data samples in $\boldsymbol{\mathcal{M}}^{*(i)}$ for local gradient computation. {We remark that Algorithm~\ref{overallAlgorithm} yields a locally optimal solution for Problem~\ref{OriginalProblem}, not a global optimum. When dealing with a non-convex optimization problem, attaining a globally optimal solution is typically not guaranteed. In such scenarios, the pursuit of a locally optimal solution becomes the conventional objective\cite{nonlinear}.}

\begin{algorithm}[ht] \small 
\caption{The Algorithm for Solving Problem~\ref{EqualProblem}}
\begin{algorithmic}[1]
    \STATE{\textbf{Input:} $N$, $K$, $B$, $L$, $T$, $U$, $N_0$, $\lambda$, $f_k$, $F_k$, $\kappa$, $q_k$, $c_k$, $|\hat{\mathcal{D}}_k|$, $\alpha_{k}^{(i)}$, $\epsilon_k$, $h_{k,n}^{(i)}$, $p_k^{\rm{max}}$, and $\sigma_{k,j}^{(i)}$}.
    \STATE {Solve Problem~\ref{ResAllProblem} via Algorithm~\ref{algresall}} to obtain $(\boldsymbol{\rho}^{*(i)}, \mathbf{p}^{*(i)})$.
    \STATE {Solve Problem~\ref{DataSelProblem} via Algorithm~\ref{algdatsel}} to obtain $\boldsymbol{\delta}^{*(i)}$.
    \STATE{\textbf{Output:} $(\boldsymbol{\delta}^{*(i)}, \boldsymbol{\rho}^{*(i)}, \mathbf{p}^{*(i)})$}.
\end{algorithmic}\label{overallAlgorithm}
\end{algorithm}

\section{Solution of Problem~\ref{ResAllProblem}}\label{sectionSolutionToProblem3}
Like Problem~\ref{EqualProblem}, Problem~\ref{ResAllProblem} is also a highly challenging mixed-integer non-convex problem. In the sequel, based on the matching theory and convex-concave procedure (CCP), we will propose a low-complexity algorithm to solve Problem~\ref{ResAllProblem}.

\subsection{Proposed Algorithm}
In this subsection, we first define a matching model. Then we give the details of the proposed matching-based algorithm. Finally, the convergence and complexity of our proposed algorithm will be analyzed.

\subsubsection{Matching Formulation}
To rationally utilize the limited communication resources, we only allocate RBs to the available devices. Let $\mathcal{U}^{(i)}$ denote the set of $U^{(i)}$ available devices. To devise a low-complexity algorithm, we consider the devices in $\mathcal{U}^{(i)}$ and RBs in $\mathcal{N}$ as two sets of nodes in a bipartite graph. Then, our goal is to match the devices to the RBs and allocate power appropriately to minimize the net cost, i.e., the objective function value of Problem \ref{ResAllProblem}. In the following, we define a matching model.

\begin{Definition}[Matching Model]\label{defmatching}
Given two disjoint sets, i.e., $\mathcal{U}^{(i)}$ and $\mathcal{N}$, a matching $\Psi$ is a mapping from the set $\mathcal{U}^{(i)}\cup \mathcal{N}$ into the set of all subsets of $\mathcal{U}^{(i)}\cup \mathcal{N}$, satisfying i) $\Psi(u) \in \mathcal{N}$, ii) $\Psi(n) \subseteq \mathcal{U}^{(i)}$, iii) $\left| \Psi(u) \right|=1$ with $|\cdot|$ denoting the size of the matching, iv) $\left| \Psi(n) \right|\le Q$, and v) $n=\Psi(u) \Leftrightarrow u=\Psi(n)$, for each $u\in\mathcal{U}^{(i)}$ and $n\in\mathcal{N}$. Here, conditions i) and iii) mean that each available device can match only one RB, and conditions ii) and iv) represent that each RB can accommodate up to $Q$ available devices.
\end{Definition}

\subsubsection{Algorithm Detail}
Motivated by the housing assignment problem in \cite{2011Peer}, we introduce the concept of swap matching into the matching formulated in Definition~\ref{defmatching} and devise a matching-based algorithm to solve Problem~\ref{ResAllProblem}.  A swapping operation means that two available devices matched with different RBs exchange their matches, while the matches of other available devices remain unchanged. The corresponding power allocation and net cost under the given RB assignment is then updated via Algorithm~\ref{algpowall}, which will be detailed in Section~\ref{secPA}.  To guarantee the reduction of the net cost, a swapping operation is approved and the matching is updated only when the net cost for the given RB assignment decreases after the swap.  The above swapping operation will be repeated several times until no swapping is further approved. The proposed matching algorithm is detailed in Algorithm~\ref{algresall} and its convergence and complexity will be analyzed in the sequel. 

\begin{algorithm}[t] \small 
\caption{Matching-based Algorithm for Solving Problem~\ref{ResAllProblem}}
\begin{algorithmic}[1]
    \STATE {Initialize the matching $\Psi$ using $\Psi_0$ and set $l \to 1$.}
    \REPEAT
        \FOR{$u\in\mathcal{U}^{(i)}$}
            \FOR{$k\in\mathcal{U}^{(i)}$}
                \IF{{$\Psi_{l-1}(k) \ne \Psi_{l-1}(u)$}}
                    \STATE {Compute and compare the cost before and after the swap using Algorithm~\ref{algpowall}.}
                    \IF{net cost decreases}
                        \STATE {Update the matching, i.e., set $\Psi_{l-1} \to \Psi_{l}$.}
                    \ENDIF
                \ENDIF
            \ENDFOR
        \ENDFOR
        \STATE {Set $l \to l+1$.}
    \UNTIL{convergence}

    \STATE {\textbf{Output} $\left(\boldsymbol{\rho}^{*(i)}, \mathbf{p}^{*(i)}\right)$}.
\end{algorithmic}\label{algresall}
\end{algorithm}

\subsubsection{Convergence}
After multiple swapping operations, the matching situation  between RBs and available devices changes as follows: $\Psi_0 \rightarrow \Psi_1 \rightarrow \Psi_2 \rightarrow \cdots$, where $\Psi_l$ with $l=1,2,\cdots$ denotes the matching at the $l$-th swapping operation and $\Psi_0$ represents the initial matching. At swapping operation $l$, the matching changes from $\Psi_{l-1}$ to $\Psi_{l}$. Let $\hat{C}_{l-1}$ and $\hat{C}_{l}$ denote the net costs at matching $\Psi_{l-1}$ and matching $\Psi_{l}$, respectively. Then, we have $\hat{C}_{l}<\hat{C}_{l-1}$, namely, the net cost decreases after the swap. Since there is a certain positive net cost to the available devices to guarantee the successful uploading of the local gradients, the net cost has a lower bound, which implies that Algorithm~\ref{algresall} will converge after a finite number of swapping operations.

\subsubsection{Complexity}Finally, we discuss the computational complexity of Algorithm~\ref{algresall}. 
For each swapping operation, we should consider all possible swapping combinations, which requires $\mathcal{O}({U^{(i)}}^2)$ operations. For each swapping attempt, we need to allocate the power, and calculate and compare the net cost before and after the swapping of the available devices under the given RB assignment. Let $\mathcal{O}(X)$ represent the complexity of the power allocation algorithm (i.e., Algorithm~\ref{algpowall}), which will be analyzed in Section~\ref{secPA}. Assume that the matching remains unchanged after $V$ swapping operations. 
Then, the computational complexity of Algorithm~\ref{algresall} is $\mathcal{O}({U^{(i)}}^2VX)$.

\subsection{Power Allocation}\label{secPA}

In line 6 of Algorithm~\ref{algresall}, the net cost is minimized by optimizing the power allocation under a given RB assignment via Algorithm~\ref{algpowall}. In this subsection, we focus on how to construct Algorithm~\ref{algpowall}. Given the RB assignment, Problem~\ref{ResAllProblem} becomes the problem in (\ref{ResAllProblem02}), shown at the top of next page.

\begin{figure*}[ht]\vspace{-8mm}
\begin{footnotesize}
\begin{align}
&\min_{\boldsymbol{p}^{(i)}}\ C^{\rm com}\left( \boldsymbol{\rho}^{*(i)},\boldsymbol{p}^{(i)} \right)+C^{\rm cmp}\label{ResAllProblem02}\\
&\mathrm{s.t.}\;
\sum_{n\in\mathcal{N}}{B\log\left(1+\frac{\rho_{k,n}^{*(i)}p_{k,n}^{(i)}h_{k,n}^{(i)}}{\sum_{t\in\mathcal{S}_n^{(i)}}{\mathbbm{1}\left[ h_{t,n}^{(i)}<h_{k,n}^{(i)} \right]\rho_{t,n}^{*(i)}p_{t,n}^{(i)}h_{t,n}^{(i)}}+N_0}\right)}T \ge \alpha_k^{(i)}L, ~{\forall}k\in \mathcal{K},
\label{consOfPowAll01}\\
&\qquad 0 \le p_{k,n}^{(i)} \le \rho_{k,n}^{*(i)}p_k^{\rm max},~{\forall}k\in \mathcal{K},~{\forall}n\in \mathcal{N}\label{maxpower02}.
\end{align}
\end{footnotesize}
\hrulefill
\end{figure*}Due to the non-convexity of the constraint in (\ref{consOfPowAll01}), the problem in (\ref{ResAllProblem02}) is a non-convex problem. To solve it in a more tractable manner, we perform some transformations as detailed below.

Specifically,  we assume that the devices in the set $\mathcal{S}_n^{(i)}$ occupying RB $n$ are arranged in ascending order according to their channel power gains. Then, the term $\sum_{t\in\mathcal{S}_n^{(i)}}\mathbbm{1}[ h_{t,n}^{(i)}<h_{k,n}^{(i)} ]\rho_{t,n}^{*(i)}p_{t,n}^{(i)}h_{t,n}^{(i)}+N_0$ can be re-expressed as s$I_{k,n}\left(\mathbf{p}^{(i)}\right) =\sum_{t=1}^{k-1}{\rho_{t,n}^{*(i)} p_{t,n}^{(i)}h_{t,n}^{(i)}}+N_0 $.
On this basis, the constraint in (\ref{consOfPowAll01}) can be rewritten as
\begin{align}
\sum_{n\in\mathcal{N}}{B\log\left(1+\frac{\rho_{k,n}^{*(i)}p_{k,n}^{(i)}h_{k,n}^{(i)}}{I_{k,n}\left( \mathbf{p}^{(i)} \right)}\right)}T \ge \alpha_k^{(i)}L, ~{\forall}k\in \mathcal{K}.\label{consOfPowAll02}
\end{align}
In addition, based on the property of the logarithmic function, (\ref{consOfPowAll02}) can be further equivalently transformed into (\ref{DCofPowAll}), shown at the top of next page.
\begin{figure*}[ht]\vspace{-6mm}
\begin{footnotesize}
\begin{flalign}\label{DCofPowAll}
&\sum_{n\in\mathcal{N}} \log\left(\rho_{k,n}^{*(i)}p_{k,n}^{(i)}h_{k,n}^{(i)}+I_{k,n}\left(\boldsymbol{p}^{(i)}\right)\right) -\sum_{n\in\mathcal{N}}\log\left(I_{k,n}\left(\boldsymbol{p}^{(i)}\right)\right) \ge \frac{\alpha_k^{(i)} L}{BT}, \ {\forall}k\in \mathcal{K}.
\end{flalign}
\end{footnotesize}
\hrulefill \vspace{-5mm}
\end{figure*}
As a result, by replacing the constraint in (\ref{consOfPowAll01}) with (\ref{DCofPowAll}), we can transform (\ref{ResAllProblem02}) to an equivalent problem as follows
\begin{align}
&\min_{\mathbf{p}^{(i)}}\ C^{\rm com}\left( \boldsymbol{\rho}^{*(i)},\mathbf{p}^{(i)} \right)+C^{\rm cmp}\label{ResAllProblem03}\\
&\mathrm{s.t.}\ (\ref{maxpower02}), (\ref{DCofPowAll}).\nonumber
\end{align}

Since the left-hand side of (\ref{DCofPowAll}) can be regarded as a difference of two concave functions, (\ref{ResAllProblem03}) is a standard difference-of-concave (DC) problem. Using the CCP method, we can achieve a stationary point of the problem in (\ref{ResAllProblem03}). The key idea of the CCP method is to linearize $\sum_{n\in\mathcal{N}} \log\left(I_{k,n}\left(\mathbf{p}^{(i)}\right)\right)$ of the constraint in (\ref{DCofPowAll}) to obtain a convex constraint, so as to generate a series of convex subproblems that approximately solves the the non-convex problem in (\ref{ResAllProblem03}). The CCP is an iterative procedure. Let $v=0,1,2,\cdots$ be the iteration index. At iteration $v$, the convex subproblem is given by the problem in (\ref{ResAllProblem04}), shown at the bottom of next page.
\begin{figure*}[hb]\vspace{-3mm}
\hrulefill
\begin{footnotesize}
\begin{align}
&\boldsymbol{p}^{(i)}(v+1)\triangleq  \arg\min_{\boldsymbol{p}}\ C^{\rm com}\left( \boldsymbol{\rho}^{*(i)},\boldsymbol{p} \right)+C^{\rm cmp}\label{ResAllProblem04}\\
&\mathrm{s.t.}\ (\ref{maxpower02}),\nonumber \\
&\sum_{n\in\mathcal{N}}{\log\left(\rho_{k,n}^{*(i)}p_{k,n}h_{k,n}^{(i)}+I_{k,n}\left(\boldsymbol{p}\right)\right)}-\sum_{n\in\mathcal{N}}{\log\left(I_{k,n}\left( \boldsymbol{p}^{(i)}(v)\right)\right)} -\sum_{n\in\mathcal{N}}{\frac{1}{I_{k,n}\left(\boldsymbol{p}^{(i)}(v)\right)} \sum_{t=1}^{k-1}{\rho_{t,n}^{*(i)}h_{t,n}^{(i)}}\left(p_{t,n}-p_{t,n}^{(i)}(v)\right)}  \ge \frac{\alpha_k^{(i)}L}{BT},~{\forall}k\in \mathcal{K}. \nonumber
\end{align}
\end{footnotesize}
\end{figure*}Here, $\mathbf{p}^{(i)}(v)$ represents a solution of the problem in (\ref{ResAllProblem04}) at iteration $v$. Since the problem in (\ref{ResAllProblem04}) is convex, its optimal solution can be achieved via CVX~\cite{cvx}.


\begin{algorithm}[!t] 
\caption{Power Allocation Algorithm}
\label{algpowall}
\begin{algorithmic}[1]
    \STATE {Require a feasible point $\mathbf{p}^{(i)}(0)$.}
    \STATE {Set $v=0$.}
    \REPEAT
        \STATE {Solve (\ref{ResAllProblem04}) using CVX to obtain $\mathbf{p}^{(i)}(v+1)$}.
        \STATE {Set $v=v+1$}.
    \UNTIL{convergence}.
\end{algorithmic}
\end{algorithm}

\begin{figure}[ht]
    \centering
    {\includegraphics[width=0.4\textwidth]{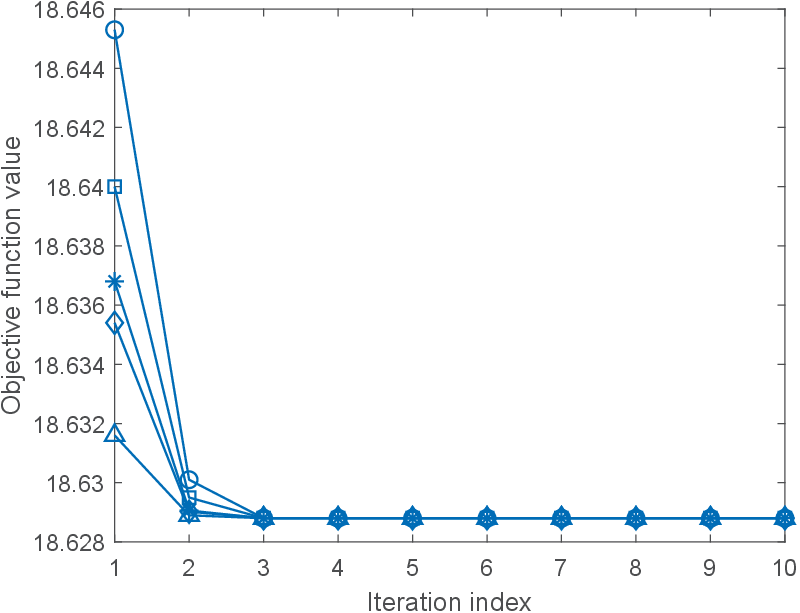}}
    \caption{The convergence curves of Algorithm~\ref{algpowall} under different initial points.}\label{figConvOfAlgPowAll}
\end{figure}

Finally, Algorithm~\ref{algpowall} details the procedure for solving the problem in (\ref{ResAllProblem03}). According to \cite{Lipp2016}, Algorithm~\ref{algpowall} converges to a stationary point of (\ref{ResAllProblem03}).
Fig.~\ref{figConvOfAlgPowAll} plots the convergence trajectory of Algorithm~\ref{algpowall} under five randomly generated initial points. From Fig.~\ref{figConvOfAlgPowAll}, we can see that
the objective function value decreases with the increase of the number of iteration index and remains unchanged after several iterations (e.g., $v\ge 4$), that is Algorithm~\ref{algpowall} converges. Furthermore, we can observe that Algorithm~\ref{algpowall} can converge to the identical objective value under different initial points, which indicates that Algorithm~\ref{algpowall} can adapt to the change of initial points robustly.
Algorithm \ref{algpowall}'s complexity is dominated by solving (\ref{ResAllProblem04}), which, therefore, can be calculated as $\mathcal{O}((KN)^{3.5}\log(1/\varepsilon))$.

\section{Solution of Problem~\ref{DataSelProblem}}\label{sectionSolutionToProblem4}
Solving Problem~\ref{DataSelProblem} optimally is NP-hard due to the existence of integer variables and the fractional form of the objective function. Toward this end, a low-complexity suboptimal algorithm for Problem~\ref{DataSelProblem} will be proposed in this section. Our algorithm consists of two stages, that is, \emph{continuous relaxation} and \emph{binary recovery}.
In the first phase, we solve the continuous relaxation of Problem \ref{DataSelProblem} via the gradient projection method. In the second stage, we achieve a feasible integer solution of Problem \ref{DataSelProblem} using the $\lambda$-representation method.\vspace{-2mm}

\subsection{Continuous Relaxation}
We first relax the integer constraint in (\ref{consOfDataSelDelta01}) to
\begin{align}
    \delta_{k,j}^{(i)}\in [0,1], {\forall}k \in \mathcal{K}, \ {\forall}j \in \mathcal{J}_k. \label{consOfDataSelDelta03}
\end{align}
Based on (\ref{consOfDataSelDelta03}), the continuous relaxation of Problem~\ref{DataSelProblem} is written as
\begin{align}
&\min_{\boldsymbol{\delta}^{(i)}}\ \lambda \hat{\Delta}(\boldsymbol{\delta}^{(i)}) + (1-\lambda) \hat{C}(\boldsymbol{\delta}^{(i)},\boldsymbol{\rho}^{*(i)}, \mathbf{p}^{*(i)})\label{relaxedDataSelProb} \\
&\mathrm{s.t.}\quad (\ref{consOfDataSelDelta02}), (\ref{consOfDataSelDelta03}).\nonumber
\end{align}

This is a non-convex continuous problem, and we can solve it efficiently by using the gradient projection method, as summarized in Algorithm~\ref{algofGPM}.
Particularly, in Step 3 of Algorithm~\ref{algofGPM}, $f(\boldsymbol{\delta}^{(i)}) $ is the objective function of the problem in (\ref{relaxedDataSelProb}), $v$ represents the iteration index, and $\alpha(v)$ denotes the diminishing stepsize at the $v$-th iteration, satisfying $\alpha(v) \to 0$ as $v\to \infty$, $\sum_{v=0}^{\infty}\alpha(v) = \infty$, and $\sum_{v=0}^{\infty}\alpha^2(v) < \infty$. In Step 4, the projection of $\hat{\boldsymbol{\delta}}^{(i)}(v+1)$ onto the feasible set of the problem in (\ref{relaxedDataSelProb}) can be achieved through tackling the following problem
\begin{align}
    {\boldsymbol{\delta}}^{(i)}(v+1) = & \arg\min_{\boldsymbol{\delta}} \left\| \boldsymbol{\delta} - \hat{\boldsymbol{\delta}}^{(i)}(v+1) \right\|^2 \label{projectOfGPM}\\
    &\mathrm{s.t.}\quad (\ref{consOfDataSelDelta02}), (\ref{consOfDataSelDelta03}).\nonumber
\end{align}The problem in (\ref{projectOfGPM}) is convex and we can achieve an optimal solution of it via CVX. Steps 3 and 4 will be repeated serval times until Algorithm~\ref{algofGPM} converges.  According to \cite{MM2017Sun}, we know that  ${\boldsymbol{\delta}}^{(i)}(v) \to {\boldsymbol{\delta}}^{\dag(i)}$ as $v\to \infty$, where ${\boldsymbol{\delta}}^{\dag(i)}$ is a stationary point of (\ref{relaxedDataSelProb}). Algorithm~\ref{algofGPM}'s complexity is dominated by solving (\ref{projectOfGPM}), which can be expressed as $\mathcal{O}((|\hat{\mathcal{D}}|)^{3.5}\log(1/\varepsilon))$.

\begin{algorithm}[ht] \small 
\caption{The Algorithm to Solve (\ref{relaxedDataSelProb})}\label{algofGPM}
\begin{algorithmic}[1]
    \STATE {Require a feasible initial point $\boldsymbol{\delta}^{(i)}{(0)}$}.
    \STATE {Set $v=0$}.
    \REPEAT
        \STATE {Compute $\hat{\boldsymbol{\delta}}^{(i)}(v+1) = \boldsymbol{\delta}^{(i)}{(v)}-\alpha(v)\nabla f(\boldsymbol{\delta}^{(i)}{(v)})$.}
        \STATE {Compute ${\boldsymbol{\delta}}^{(i)}(v+1)$ by projecting $\hat{\boldsymbol{\delta}}^{(i)}(v+1)$ onto the feasible set of the problem in (\ref{relaxedDataSelProb})}.
        \STATE {Set $v=v+1$}.
    \UNTIL{convergence}
\end{algorithmic}
\end{algorithm}

\subsection{Binary Recovery}
The obtained solution $\boldsymbol{\delta}^{\dag(i)}$ is generally continuous and hence is an infeasible solution of Problem \ref{DataSelProblem}. In the sequel, based on $\boldsymbol{\delta}^{\dag(i)}$, we construct a feasible solution of Problem~\ref{DataSelProblem}.  Let $\mathcal{B}=\{(\ref{consOfDataSelDelta01}),(\ref{consOfDataSelDelta02})\}$ be the constraint set. By projecting $\boldsymbol{\delta}^{\dag(i)}$ onto $\mathcal{B}$, we establish the following problem
\begin{align}
    \boldsymbol{\delta}^{*(i)} \triangleq \arg\min_{\boldsymbol{\delta}^{(i)}\in\mathcal{B}} \left\| \boldsymbol{\delta}^{(i)} - \boldsymbol{\delta}^{\dag(i)} \right\|^2.\label{feasibleproject}
\end{align}
The problem in (\ref{feasibleproject}) is an integer nonlinear programming. By using the $\lambda$-representation technique \cite{lambda2020Jiang}, we can transform it into a linear programming problem as follows.
\begin{align}
 &\min_{\boldsymbol{\delta}^{(i)},\boldsymbol{a},\boldsymbol{b}} \sum_{k\in\mathcal{K}}{\sum_{j\in\mathcal{J}_k}}{\left[ (\delta_{k,j}^{\dag(i)})^2a_{k,j}+(1-\delta_{k,j}^{\dag (i)})^2b_{k,j} \right]} \label{lambda-representation} \\
    & \mathrm{s.t.}\ (\ref{consOfDataSelDelta02}), (\ref{consOfDataSelDelta03}),\nonumber \\
    & \quad b_{k,j} = \delta_{k,j}^{(i)}, \ {\forall}k\in \mathcal{K},\ {\forall}j\in \mathcal{J}_k, \label{consOfrepresent01}\\
    & \quad a_{k,j} + b_{k,j} = 1, \ {\forall}k\in\mathcal{K},\ {\forall}j\in\mathcal{J}_k,\label{consOfrepresent02} \\
    & \quad a_{k,j} \ge 0,\ b_{k,j} \ge 0, \ {\forall}k\in\mathcal{K},\ {\forall}j\in\mathcal{J}_k,\label{consOfrepresent03}
\end{align}
where $\boldsymbol{a}=(a_{k,j})_{k\in\mathcal{K},j\in\mathcal{J}_k}$ and $\boldsymbol{b}=(b_{k,j})_{k\in\mathcal{K},j\in\mathcal{J}_k}$. Lemma \ref{lemmaOfRepresentation} indicates the relationship between the problems in (\ref{lambda-representation}) and  (\ref{feasibleproject}).

\begin{Lemma}\label{lemmaOfRepresentation}
The problems in (\ref{lambda-representation}) and (\ref{feasibleproject}) are equivalent.
\end{Lemma}


Please refer to Appendix \ref{proofOfLemma5} for the proof. Based on Lemma~\ref{lemmaOfRepresentation}, we know that to solve the problem in (\ref{feasibleproject}), it is only required to solve the linear problem in (\ref{lambda-representation}) by using CVX. Thus, the corresponding computational complexity for solving the problem in (\ref{lambda-representation}) can be expressed as $\mathcal{O}((3|\hat{\mathcal{D}}|)^{3.5}\log(1/\varepsilon))$.

\subsection{Algorithm Summary}
Finally, Algorithm~\ref{algdatsel} summarizes the details of solving Problem~\ref{DataSelProblem}, whose complexity is the summation of the complexity of the \emph{continuous relaxation} stage and the \emph{binary recovery} stage, which is given by $\mathcal{O}((1 +3^{3.5})|\hat{\mathcal{D}}|^{3.5}\log(1/\varepsilon))$. In summary, the complexity of Algorithm~\ref{overallAlgorithm} is given by $\mathcal{O}(U^{(i)^2}V(KN)^{3.5}\log(1/\varepsilon)+(1+3^{3.5})|\hat{\mathcal{D}}|^{3.5}\log (1/\varepsilon))$.

\begin{algorithm}[ht] \small 
\caption{The Algorithm to Solve Problem~\ref{DataSelProblem}}\label{algdatsel}
\begin{algorithmic}[1]
    \STATE {Obtain $\boldsymbol{\delta}^{\dag(i)}$ using Algorithm~\ref{algofGPM}}.
    \STATE {Obtain $\boldsymbol{\delta}^{*(i)}$ by solving the problem in (\ref{feasibleproject}) via CVX}.
    \STATE {\textbf{Output:} $\boldsymbol{\delta}^{*(i)}$.}
\end{algorithmic}
\end{algorithm}

\section{Simulation Results}\label{sectionSimuRes}
Extensive simulations will be conducted in this section to show the superiority of our proposed scheme (obtained by Algorithm~\ref{overallAlgorithm}). In the following, we will first introduce the simulation setup and then detail the performance evaluation.

\subsection{Simulation Setup}
Unless otherwise stated, the simulation parameters are set as follows. We consider that the FEEL system has $K=10$ devices. For device $k$, we set the cost per Joule and the reward for each data sample, respectively, to $c_k=5$ and $q_k=0.002$, if $k$ is odd, and $c_k=10$ and $q_k=0.005$, otherwise. The CPU frequency, the number of CPU cycles to handle a sample, and the capacitance coefficient are set to $f_k=\{0.1, 0.2, \cdots, 1.0\}$ GHz, $F_k=20$ cycles/sample, and $\kappa=1\times 10^{-28}$, respectively.
The probability of local gradient uploading is $\epsilon_k=0.2$ if $k$ is odd, and $\epsilon_k=0.8$, otherwise \cite{resall2022Wen}. {The maximum power limit of device $k$ is set to $p_k^{\rm{max}}=10$~W {\cite{cao2021optimized}}.} In addition, we set $N=5$, $Q=2$, $B=2$ MHz, $N_0=10^{-9}$ W, $T=500$ ms, $\lambda=1\times10^{-3}$. The channel power gain $h_{k,n}$ follows an exponential distribution with mean $10^{-5}$.

We train an image classification model with the proposed FEEL algorithm. The corresponding datasets are MNIST and Fashion-MNIST. The MNIST dataset has $70000$ handwritten grayscale images of the digits 0 to 9, and the Fashion-MNIST dataset comprises $70000$ grayscale images of fashion items from ten classes, e.g., ``t-shirt'', ``pants'', and ``bag''. {For each dataset, $60000$ images are used for model training and the rest are the test dataset.} To simulate non-IID distribution, we randomly allocate $|\mathcal{D}_k|=1000$ figures of one label to device $k$ in set $\mathcal{K}$, and then randomly choose $|\hat{\mathcal{D}}_k|=200$ samples from $\mathcal{D}_k$ in each communication round. {We assume that some data samples on each device are mislabeled.
Let $\varrho_k$ denote the proportion of mislabeled data on device $k$. To simulate the scenario of data mislabeling, we randomly select $\varrho_k\cdot\left|\mathcal{D}_k\right|$ of the data samples on device $k$ and then mislabel each of them, such as labeling the digit ``1'' as ``0'', or labeling the item ``t-shirt'' as ``pants''. In the simulations, we set the mislabeled proportion $\varrho_k$ to $10\%$ for all $k\in\mathcal{K}$.}  We apply the convolutional neural network (CNN) to conduct the image classification task. The CNN model has seven layers, two~$5\times5$ convolution layers (the first and second layers have 10 and 20 channels, respectively, each followed by $2\times2$ max pooling), and three full connection layers with ReLu activation.
Then using Python, the size of the gradient $\mathbf{\hat{g}}_k^n$ is estimated to be $ \ell = 0.56 \times 10^6$ bits (MNIST) or $ \ell = 1\times 10^6$ bits (Fashion-MNIST). We choose Adam as the optimizer and set the learning rate $\eta$ to~$0.001$. {Please note that the hyper-parameters employed in this study have been selected based on prior experimental experience and analogous experimental configurations, rather than through a process of fine-tuning.}

To demonstrate the superior of our proposed scheme, the following four representative baseline schemes are investigated { \cite{joint2020yu,resall2022Wen,data-SGD2020He}}:
\begin{itemize}
  \item Baseline 1: Device $k$ randomly selects half of the local data samples for gradient computing and uploads the gradient using one RB with minimal channel power gain.
  \item Baseline 2: Device $k$ randomly selects half of the local data samples for gradient computing and uploads the gradient using one RB with maximal channel power gain.
  \item Baseline 3: Device $k$ selects all of the local data samples for gradient computing and uploads the gradient using one RB with minimal channel power gain.
  \item Baseline 4: Device $k$ selects all of the local data samples for gradient computing and uploads the gradient using one RB with maximal channel power gain.
\end{itemize}Note that the power allocation of the four baseline schemes can be achieved via Algorithm~\ref{algpowall}.

\subsection{Evaluation Details}

\begin{figure}[ht]\vspace{-5mm}
    \centering
    \subfloat[Convergence on the MNIST dataset.]{
    \includegraphics[width=0.23\textwidth]{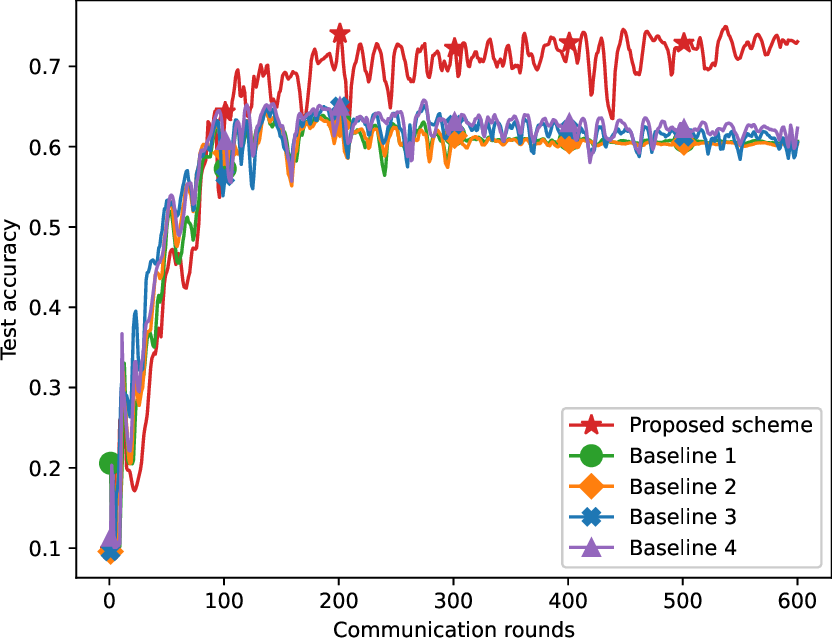}}
    \subfloat[Cumulative net cost on the MNIST dataset.]{
    \includegraphics[width=0.23\textwidth]{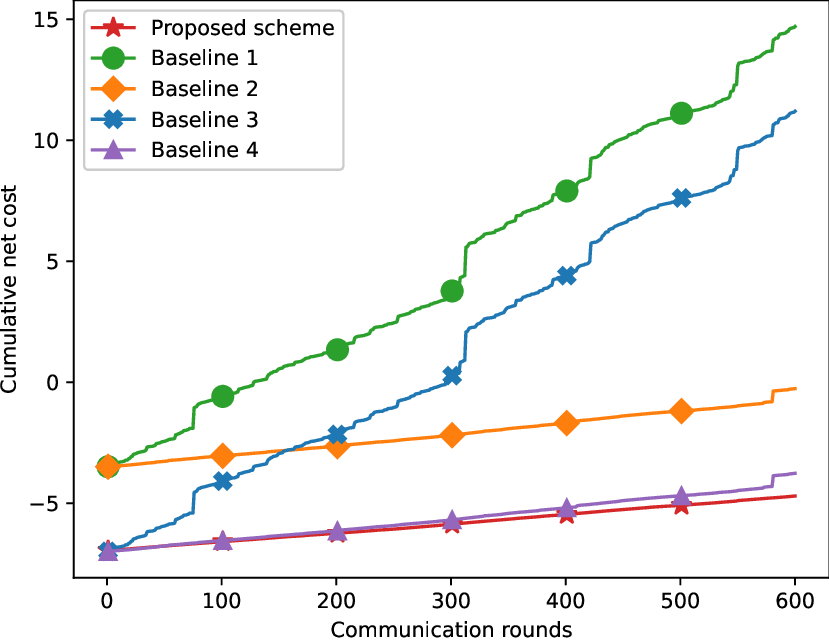}}\\
    \subfloat[Convergence on the Fashion-MNIST dataset.]{\includegraphics[width=0.23\textwidth]{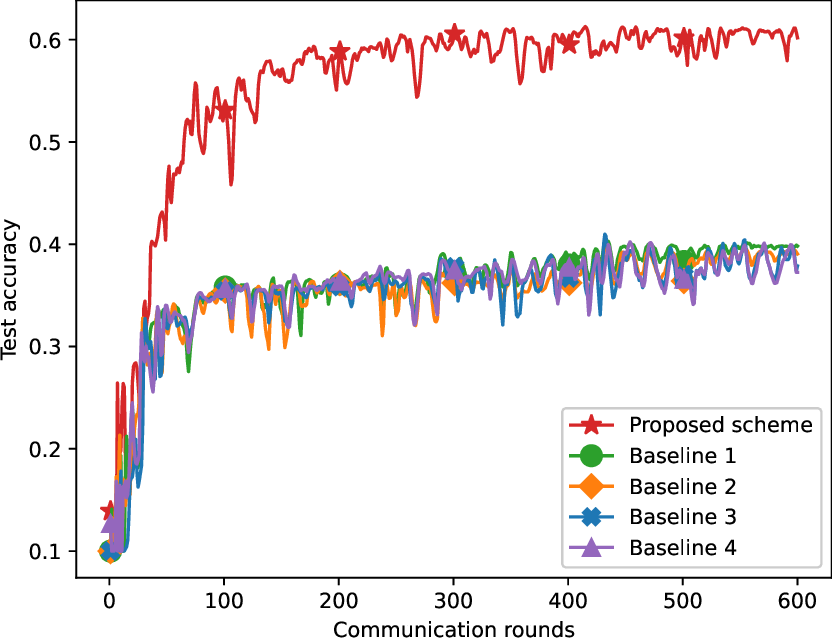}}\hspace{1mm}
    \subfloat[Cumulative net cost on the Fashion-MNIST dataset.]{\includegraphics[width=0.23\textwidth]{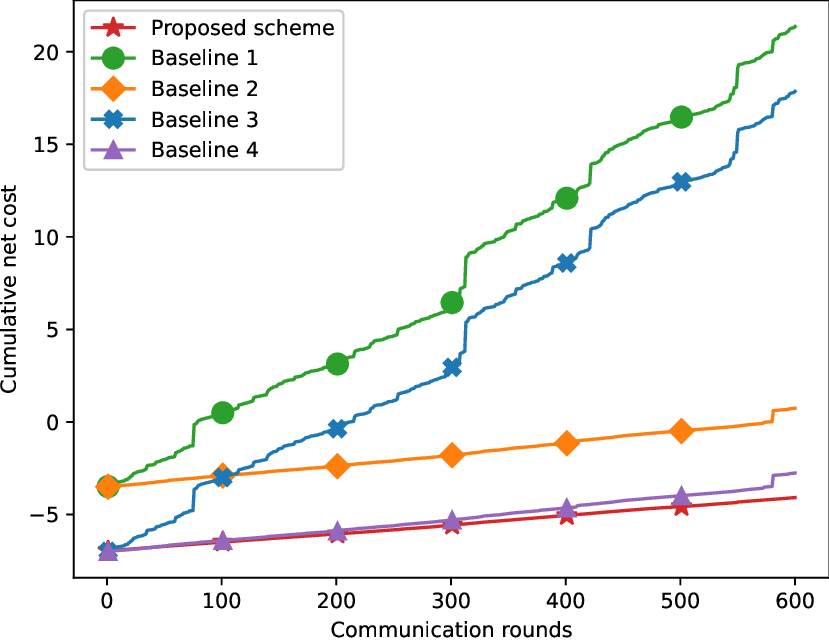}}
    \caption{Convergence and cumulative net cost of FEEL under different schemes.}\label{figComMNIST}
    \vspace{-2mm}  
\end{figure}
Fig.~\ref{figComMNIST} depicts the convergence of model training and the \textit{cumulative} net cost under the proposed scheme and the baseline schemes. Here, the cumulative net cost at communication round $i$ refers to the sum of the net costs of the first $i$ rounds.
From Fig.~\ref{figComMNIST}, we have some observations.
First, the test accuracy of all schemes increases in a fluctuating manner as the iteration goes on because the local and global models are trained on the user devices and the edge server, respectively.
Second, as the iteration proceeds, the proposed scheme can achieve significant performance gain in terms of test accuracy and cumulative net cost compared to the four baselines. For example, at the $600$th communication round, the test accuracy of the proposed scheme on the MNIST dataset ({\textit{resp.}} Fashion-MNIST dataset) can be improved by about 20\% ({\textit{resp.}} 51\%), but the cumulative net cost is reduced by at least 19\% ({\textit{resp.}} 32\%). Such performance gain stems from the fact that the proposed scheme can wisely select training data samples and appropriately allocate radio resources at each communication round.  Note that the negative cumulative net costs in Fig.~\ref{figComMNIST} indicate that users are profitable under these schemes, which can encourage them to participate in model training more actively.
Finally, we find that the difference in test accuracy between the four baseline schemes is not so significant, indicating that if the mislabeled data samples of each device are not excluded during model training, FEEL can only maintain a relatively low level of test accuracy. The reason is that the mislabeled data samples will produce incorrect gradient information, which will mislead the convergence behavior of FEEL.

\begin{figure}[ht]\vspace{-5mm}
    \centering
    \subfloat[Test accuracy on the MNIST dataset.]{
    \includegraphics[width=0.23\textwidth]{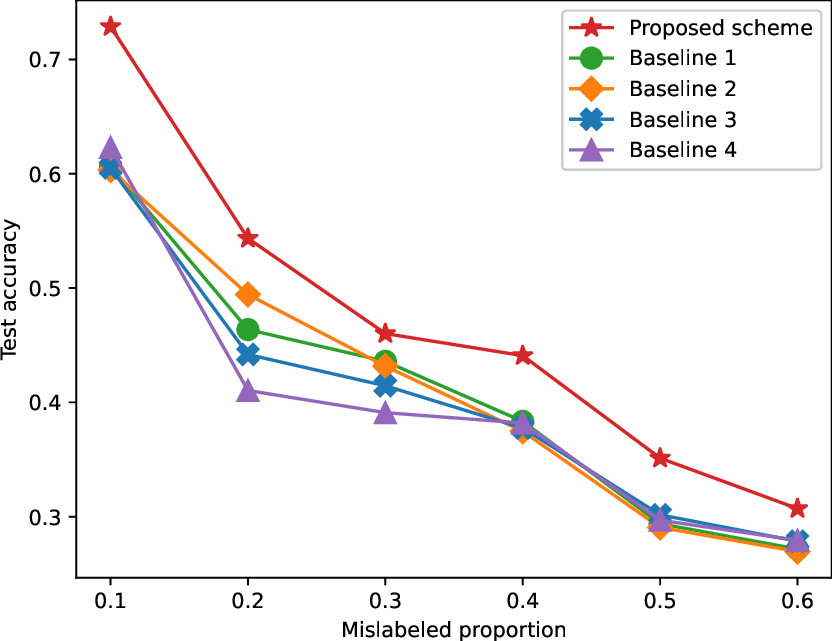}} 
    \subfloat[Cumulative net cost on the MNIST dataset.]{
    \includegraphics[width=0.23\textwidth]{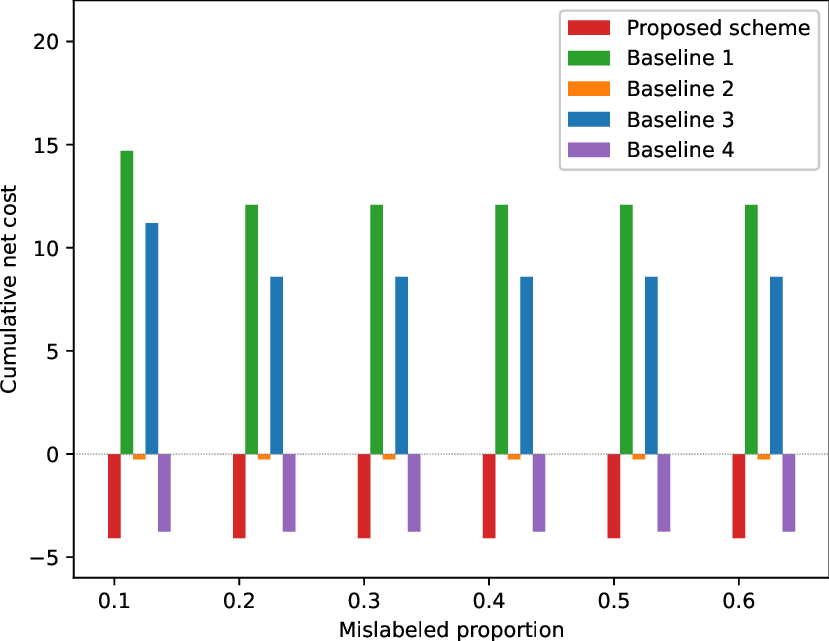}}\\
    \subfloat[Test accuracy on the Fashion-MNIST dataset.]{\includegraphics[width=0.23\textwidth]{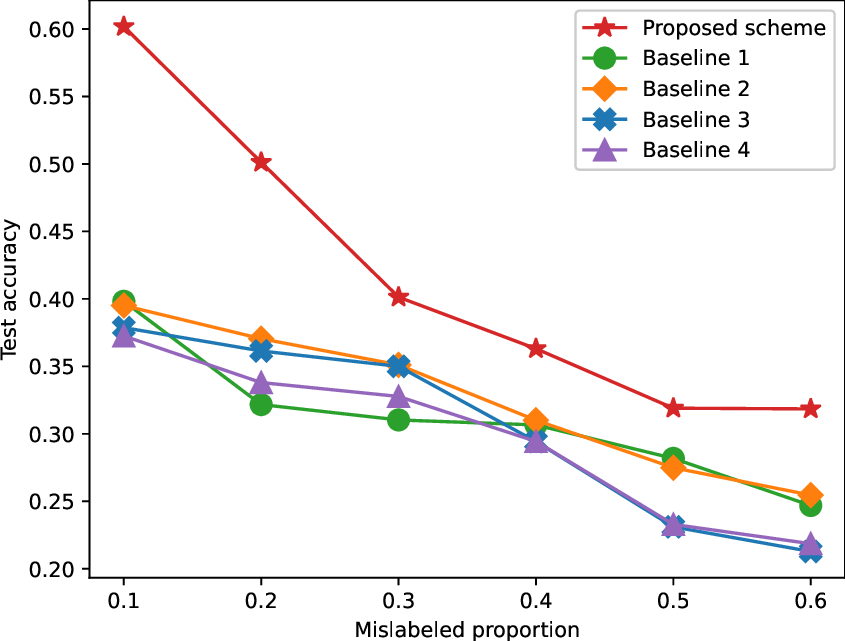}}\hspace{1mm}
    \subfloat[Cumulative net cost on the Fashion-MNIST dataset.] {\includegraphics[width=0.23\textwidth]{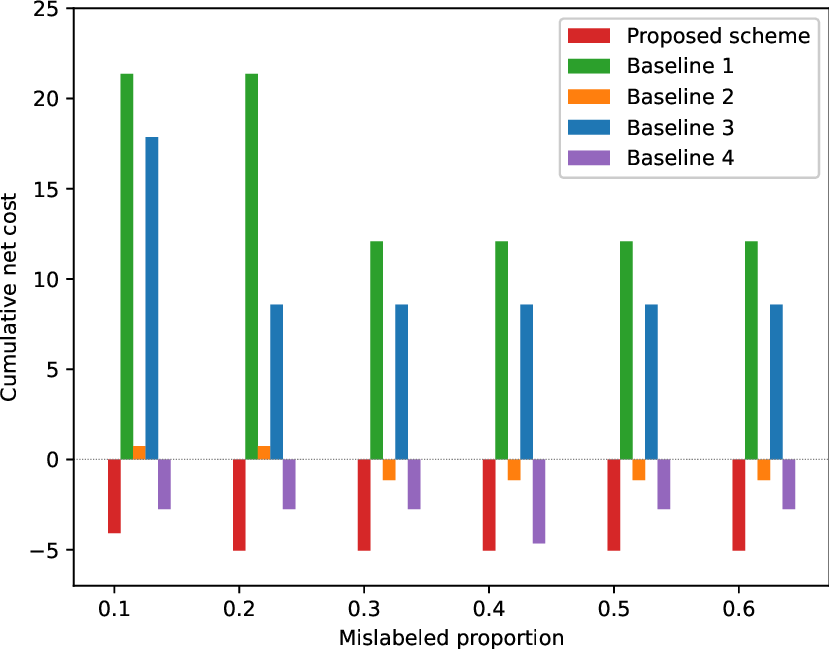}}
    \caption{ Effect of the mislabeled proportion.}\label{figComRatio}
    \vspace{-2mm}  
\end{figure}

Fig.~\ref{figComRatio} shows the effect of the mislabeled proportion under the proposed scheme and the baseline schemes.
All results in Fig.~\ref{figComRatio} are obtained when the training process of FEEL stops at the $300$th communication round.
As expected, we can see that the test accuracy under all schemes significantly decreases as the mislabeled proportion increases, but this does not hold for the net cost which is independent of the mislabeled proportion (see Problem~\ref{OriginalProblem}).
In addition, we can also see that our solution is dramatically better than the baseline schemes, indicating that the proposed scheme is more robust than the baseline schemes in combating data~mislabeling.

\begin{figure}[ht]\vspace{-5mm}
    \centering
    \subfloat[Test accuracy on the MNIST dataset.]{\includegraphics[width=0.23\textwidth]{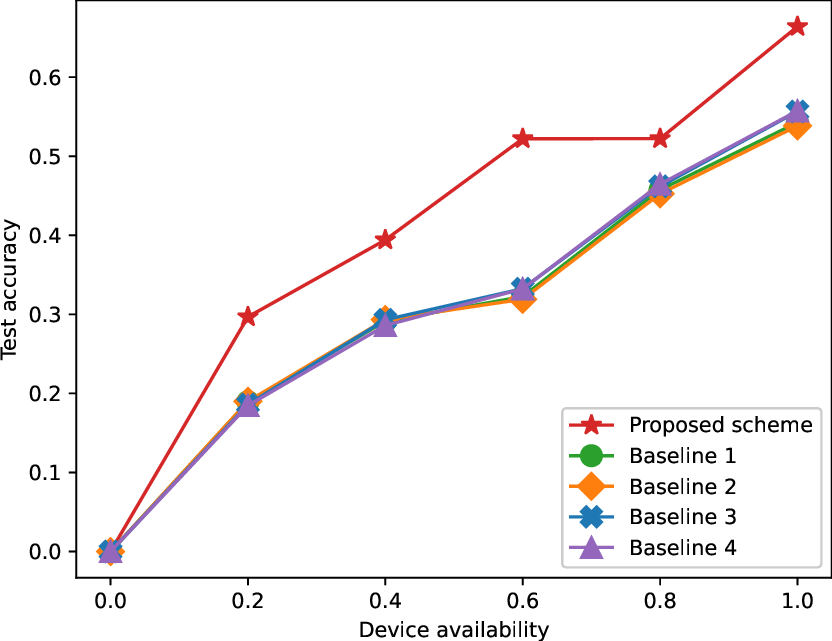}}
    \subfloat[Cumulative net cost on the MNIST dataset.]{\includegraphics[width=0.23\textwidth]{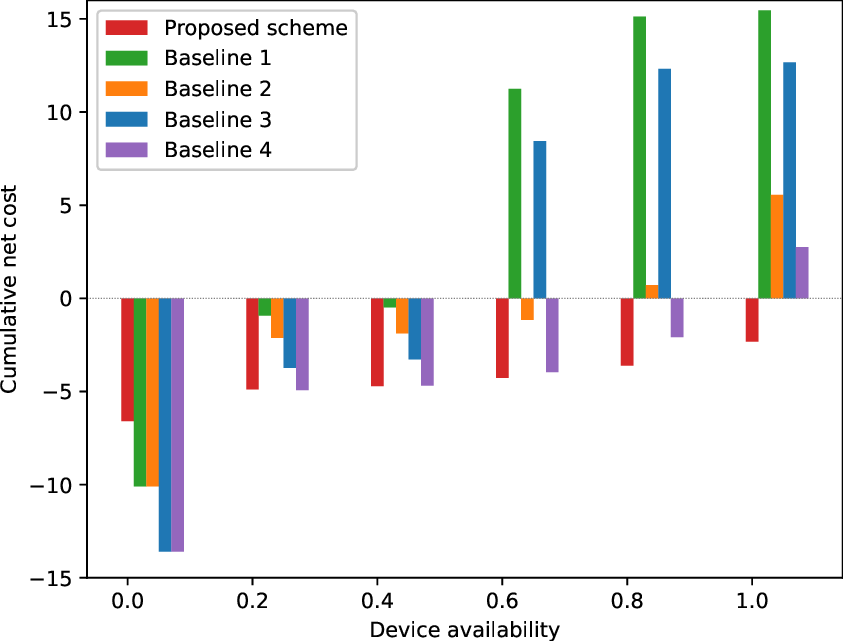}}\\
    \subfloat[Test accuracy on the Fashion-MNIST dataset.]{\includegraphics[width=0.23\textwidth]{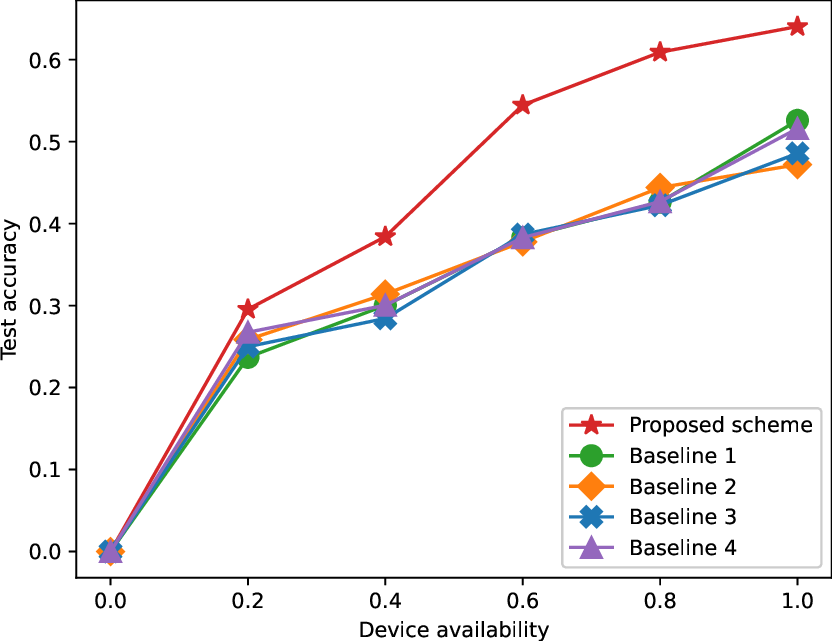}}
    \subfloat[Cumulative net cost on the Fashion-MNIST dataset.]{\includegraphics[width=0.23\textwidth]{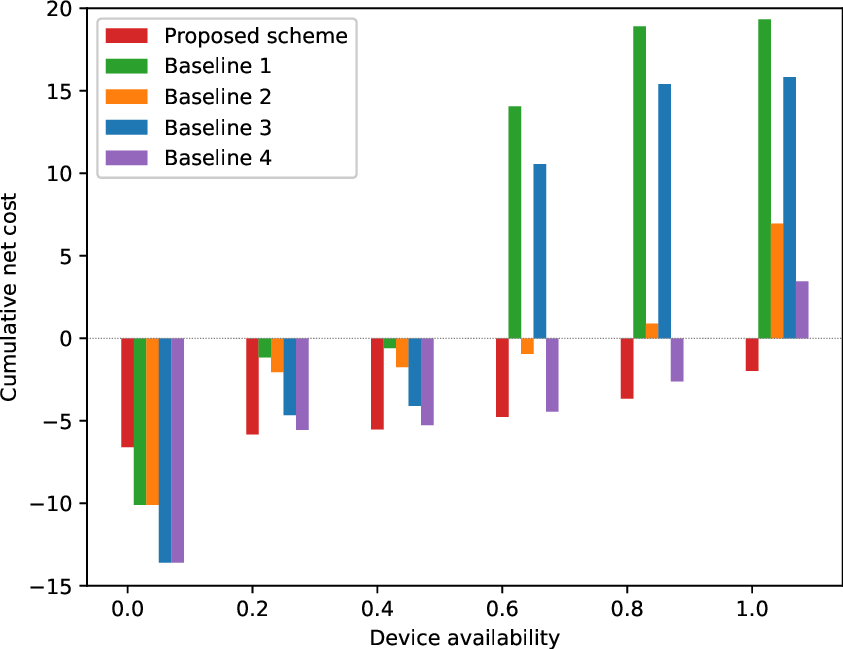}}
    \caption{Effect of the device availability.}\label{figComAvaiRatio}
    \vspace{-2mm}
\end{figure}

Fig.~\ref{figComAvaiRatio} shows the effect of the device availability under the proposed scheme and the baseline schemes. All results in Fig.~\ref{figComAvaiRatio} are obtained when the training process of FEEL stops at the $300$th communication round.
From Fig.~\ref{figComAvaiRatio}, we can see that the test accuracy improves with the device availability, but the more devices participate in local gradient uploading, the greater the cumulative cost. Note that when all devices are unavailable to upload gradients, i.e., the device availability $\epsilon_k=0$ for all $k\in\mathcal{K}$, the test accuracy under each scheme is zero since the server cannot aggregate any gradients from the devices. Our scheme in the case of $\epsilon_k=0$ may underperform the baseline schemes because the baselines can contribute more data samples than ours. However, when some devices become available for gradient uploading (e.g., when $\epsilon_k \ge 0.2$), the proposed scheme becomes superior to the baselines, indicating that our solution can better adapt to the changes in device availability.

\section{Conclusions}\label{conclusions}
In this paper, we first rigorously model the training process of FEEL and derive its one-round convergence bound. Then, we formulate a joint resource allocation and data selection optimization problem, which, unfortunately, cannot be solved directly. To tackle this problem, we equivalently transform it into a more tractable form with some appropriate transformations and then break it into the resource allocation problem and the data selection problem.  Both subproblems are mixed-integer non-convex and integer non-convex optimization problems, respectively, and it is very challenging to obtain their optimal solutions. Based on the matching theory and applying the convex-concave procedure and gradient projection methods, we propose a low-complexity suboptimal algorithm for the resource allocation problem and the data selection problem, respectively. At last, the superiority of the proposed scheme is demonstrated via extensive numerical results. {In future work, we aim to extend the results proposed in this paper to multi-task FEEL systems. Specifically, we will explore how machine learning methods can be leveraged to achieve cross-domain data selection within this system. By doing so, we seek to enhance the system's adaptability and efficiency across different domains while ensuring that data selection remains optimized for each specific task.}

\begin{appendices}\vspace{-5mm}
\section{Proof of Lemma~\ref{lemmaOfUnbias}}\label{proofOfLemma1}
The derivative of $L(\mathbf{w})$ at $\mathbf{w}=\mathbf{w}^{(i)}$ is calculated as
\begin{align}
    \mathbf{g}^{(i)}&=\nabla L\left(\mathbf{w}^{(i)}\right) \label{gndglbgrad}\nonumber \\
    &=\frac{1}{|\mathcal{D}|}\sum_{k\in\mathcal{K}}{|\mathcal{D}_k|\nabla L_k\left(\mathbf{w}^{(i)}\right)} \nonumber \\
    &=\frac{1}{|\mathcal{D}|}\sum_{k\in\mathcal{K}}{|\mathcal{D}_k|\mathbf{g}_k^{(i)}}.
\end{align}
Here, $\mathbf{g}_k^{(i)}\triangleq \nabla L_k(\mathbf{w}^{(i)})= \frac{1}{|\mathcal{D}_k|}\sum_{(\mathbf{x}_j,y_j) \in \mathcal{D}_k}{\nabla \ell(\mathbf{w}^{(i)},\mathbf{x}_j,y_j)}$ denotes the ground-truth local gradient of the $k$-th device.

{\color{black}The expectation of $\mathbf{\hat{g}}^{(i)}$ is calculated as
\begin{align}
    \mathbb{E}\left[ \mathbf{\hat{g}}^{(i)} \right] &=\mathbb{E}\left[ \frac{1}{|\hat{\mathcal{D}}|}\sum_{k\in\mathcal{K}}{\frac{|\hat{\mathcal{D}}_k|}{\epsilon _k} \alpha _k^{(i)}\mathbf{\hat{g}}_k^{(i)}} \right] \label{expofglobalgrad}\nonumber \\
    &\mathop = \limits^{(\mathrm{a})}\frac{1}{|\hat{\mathcal{D}}|}\sum_{k\in\mathcal{K}}{|\hat{\mathcal{D}}_k| \mathbb{E}\left[ \frac{\alpha _k^{(i)}}{\epsilon _k} \right]  \mathbb{E}\left[ \mathbf{\hat{g}}_k^{(i)} \right]}\nonumber \\
    &\mathop = \limits^{(\mathrm{b})}\frac{1}{|\hat{\mathcal{D}}|}\sum_{k\in\mathcal{K}}{|\hat{\mathcal{D}}_k|  \mathbb{E}\left[ \mathbf{\hat{g}}_k^{(i)} \right] }.
\end{align}Here, (a) follows from the independence between the random variables $\alpha _k^{(i)}$ and $\mathbf{\hat{g}}_k^{(i)}$ and (b) follows from $\mathbb{E}[\alpha _k^{(i)}] = \epsilon _k$. According to~\cite{joint2020yu}, $\mathbf{\hat{g}}_k^{(i)}$ is unbiased, making their average in (\ref{expofglobalgrad}) to be an unbiased estimation of $\mathbf{g}^{(i)}$, which completes the proof.}

\section{Proof of Lemma~\ref{lemmaOfConvRate}}\label{proofOfLemma2}
First, based on the Lipschitz continuity of $\nabla L(\mathbf{w})$, we obtain
\begin{align}\label{lipschitz}
    L(\mathbf{u}) \le L(\mathbf{v}) +(\nabla L(\mathbf{v})) ^{\rm{T}}(\mathbf{u}-\mathbf{v}) +\frac{\beta}{2}\| \mathbf{u}-\mathbf{v} \| ^2,
\end{align}
where $\beta>0$ is a modulus, $(\cdot)^{\rm{T}}$ denotes the transposition operator. Based on (\ref{lipschitz}), we have the following inequality:
\begin{align}\label{lipschitz02}
    &L\left(\mathbf{w}^{(i+1)}\right) \le L\left(\mathbf{w}^{(i)}\right)+\left(\nabla L\left(\mathbf{w}^{(i)}\right)\right)^{\rm{T}}\left(-\eta^{(i)}\hat{\mathbf{g}}^{(i)}\right)\nonumber \\
    &+\frac{\beta}{2}\left\| -\eta^{(i)} \hat{\mathbf{g}}^{(i)} \right\|^2 \nonumber\\
    &=L\left(\mathbf{w}^{(i)}\right)-\eta^{(i)} \left(\mathbf{g}^{(i)}\right)^{\rm{T}}\hat{\mathbf{g}}^{(i)}+\frac{\beta}{2}\left(\eta^{(i)}\right)^2 \left\| \hat{\mathbf{g}}^{(i)} \right\|^2.
\end{align}

Next, by taking the expectation on both sides of~(\ref{lipschitz02}), we can obtain
\begin{footnotesize}\begin{align}\label{convBound02}
    &\mathbb{E}\left[ L\left(\mathbf{w}^{(i+1)}\right) -L(\mathbf{w}^*)\right]
    \le \mathbb{E}\left[ L\left(\mathbf{w}^{(i)}\right) -L(\mathbf{w}^*)\right] \nonumber \\
    & - \eta^{(i)} \left(\mathbf{g}^{(i)}\right) ^{\rm{T}}\mathbb{E}\left[ \mathbf{\hat{g}}^{(i)} \right] + \frac{\beta \left(\eta^{(i)}\right) ^2}{2}\mathbb{E}\left[\left\|\mathbf{\hat{g}}^{(i)}\right\|^2 \right] \nonumber \\
    &\mathop = \limits^{(\mathrm{a})} \mathbb{E}\left[ L\left(\mathbf{w}^{(i)}\right) -L(\mathbf{w}^*)\right] -\eta^{(i)} \left\| \mathbf{g}^{(i)} \right\| ^2+\frac{\beta \left(\eta^{(i)}\right) ^2}{2}\mathbb{E}\left[ \left\| \hat{\mathbf{g}}^{(i)} \right\|^2 \right],
\end{align}\end{footnotesize} where (a) is due to the unbiasedness of $\hat{\mathbf{g}}^{(i)}$ (see Lemma~\ref{lemmaOfUnbias}).

Finally, by applying the generalized triangle inequality $\lVert \sum_{j=1}^{n}{x_j} \rVert^2 \le n\sum_{j=1}^n{\lVert x_j \rVert^2}$ \cite{gentri2008Hsu}, we further have
\begin{footnotesize}\begin{align}
&\mathbb{E}\left[\left\| \hat{\mathbf{g}}^{(i)} \right\|^2\right]=\mathbb{E}\left[\left\| \frac{1}{|\mathcal{D}|}\sum_{k\in\mathcal{K}}{\frac{\alpha_k^{(i)}|\mathcal{D}_k|}{\epsilon_k}\hat{\mathbf{g}}_k^{(i)}} \right\|^2\right] \nonumber\\
&\le \mathbb{E}\left[\frac{1}{|\mathcal{D}|^2}\left(\sum_{k\in\mathcal{K}}{\frac{\alpha_k^{(i)}|\mathcal{D}_k|}{\epsilon_k}}\right)\left(\sum_{k\in\mathcal{K}}{\frac{\alpha_k^{(i)}|\mathcal{D}_k|}{\epsilon_k}}\left\| \hat{\mathbf{g}}_k^{(i)} \right\|^2\right)\right]\nonumber\\
& \le  \frac{1}{|\mathcal{D}|^2}\mathbb{E}\left[\left(\sum_{k\in\mathcal{K}}{\frac{\alpha_k^{(i)}|\mathcal{D}_k|}{\epsilon_k}}\right)\left(\sum_{k\in\mathcal{K}}{\frac{\alpha_k^{(i)}|\mathcal{D}_k|}{\epsilon_k|\mathcal{M}_k^{(i)}|}\sum_{(\mathbf{x}_j,y_j) \in \mathcal{M}_{k}^{(i)}}{\left\| g_{k,j}^{(i)} \right\|^2}}\right)\right] \nonumber \\
&=\frac{1}{|\mathcal{D}|^2}\sum_{k\in \mathcal{K}}{\Delta(\boldsymbol{\mathcal{M}}^{(i)})},\label{ineqEghat}
\end{align}\end{footnotesize}where $\Delta(\boldsymbol{\mathcal{M}}^{(i)})$ is given in (\ref{expOfDelta}). Substituting (\ref{ineqEghat}) into (\ref{convBound02}), we obtain Lemma \ref{lemmaOfConvRate}.

{\color{black}\section{Proof of Lemma~\ref{lemmaOfConvBnd}}\label{proofOfLemma3}
Since $L(\mathbf{w})$ is strongly convex, we obtain
\begin{align}\label{StrongConv}
    L(\mathbf{w}^{(i+1)}) &\ge L(\mathbf{w}^{(i)})+\left(\mathbf{g}^{(i)}\right)^T\left(\mathbf{w}^{(i+1)}-\mathbf{w}^{(i)}\right)\nonumber \\
    &+\frac{\mu}{2}\|\mathbf{w}^{(i+1)}-\mathbf{w}^{(i)}\|^2.
\end{align}Minimizing both sides of (\ref{StrongConv}) with respect to $\mathbf{w}^{(i+1)}$, it follows
\begin{align}\label{minStrongConv}
    \min_{\mathbf{w}^{(i+1)}}L(\mathbf{w}^{(i+1)}) &\ge
    \min_{\mathbf{w}^{(i+1)}}[L(\mathbf{w}^{(i)})+\left(\mathbf{g}^{(i)}\right)^T\left(\mathbf{w}^{(i+1)}-\mathbf{w}^{(i)}\right)\nonumber \\
    &+\frac{\mu}{2}\|\mathbf{w}^{(i+1)}-\mathbf{w}^{(i)}\|^2].
\end{align}The minimization of the left-hand side of (\ref{minStrongConv}) is achieved when $\mathbf{w}^{(i+1)}=\mathbf{w}^*$, while the right-hand side of (\ref{minStrongConv}) is minimized when $\mathbf{w}^{(i+1)}=\mathbf{w}^{(i)}-\frac{1}{\mu}\mathbf{g}^{(i)}$. Based on this, we have
\begin{align}\label{graStrongConv}
\|\mathbf{g}^{(i)}\|^2 \ge 2\mu \left(L(\mathbf{w}^{(i)})-L(\mathbf{w}^*)\right).
\end{align}

Combining (\ref{minStrongConv}) with the one-round convergence rate in (\ref{convBound}), we obtain
\begin{footnotesize}\begin{align}\label{ConvBnd02}
&\mathbb{E}\{L(\mathbf{w}^{(i+1)})-L(\mathbf{w}^*)\}\le \mathbb{E}\{L(\mathbf{w}^{(i)})-L(\mathbf{w}^*)\} \nonumber \\
&-2\mu\eta^{(i)}
\mathbb{E}\{L(\mathbf{w}^{(i)})-L(\mathbf{w}^*)\}+\frac{\beta(\eta^{(i)})^2}{2|\hat{\mathcal{D}}|^2}\Delta(\boldsymbol{\mathcal{M}}^{(i)})\nonumber\\
&=(1-2\mu\eta^{(i)})\mathbb{E}\{L(\mathbf{w}^{(i)})-L(\mathbf{w}^*)\}
+\frac{\beta(\eta^{(i)})^2}{2|\hat{\mathcal{D}}|^2}\Delta(\boldsymbol{\mathcal{M}}^{(i)})\nonumber\\
&\mathop \le \limits^{(\mathrm{recursively})} \prod \limits_{t=1}^i(1-2\mu\eta^{(t)})\mathbb{E}\{L(\mathbf{w}^{(1)})-L(\mathbf{w}^*)\}\nonumber \\
&+\frac{\beta}{2|\hat{\mathcal{D}}|^2}\sum_{t=1}^i{A^{(t)}(\eta^{(t)})^2\Delta(\boldsymbol{\mathcal{M}}^{(t)})},
\end{align}\end{footnotesize}with $A^{(t)}=\prod \limits_{j=t+1}^i (1-2\mu \eta^{(j)})$, we obtain Lemma~\ref{lemmaOfConvBnd}.
}

\section{Proof of Lemma~\ref{lemmaOfRepresentation}}\label{proofOfLemma5}
We first point out that the term of $\left\| \boldsymbol{\delta}^{(i)} - \boldsymbol{\delta}^{\dag(i)} \right\|^2$ can be represented as $\left\| \boldsymbol{\delta}^{(i)} - \boldsymbol{\delta}^{\dag(i)} \right\|^2=\sum_{k\in\mathcal{K}}{\sum_{j\in\mathcal{J}_k}{\left( \delta_{k,j}^{(i)}-\delta_{k,j}^{\dag(i)} \right)^2}}$. Based on the $\lambda$-representation technique \cite{ILPtoLP1977}, the integer convex function ${\left( \delta_{k,j}^{(i)}-\delta_{k,j}^{\dag(i)} \right)^2}$ is equivalent to the following linear problem
\begin{align}
    & \min_{a_{k,j},b_{k,j}}\left( \delta_{k,j}^{\dag(i)} \right)^2a_{k,j}+\left( 1-\delta_{k,j}^{\dag(i)} \right)^2b_{k,j}  \label{lamrepresent01} \\
    & \mathrm{s.t.} \ b_{k,j} = \delta_{k,j},\nonumber\\
    & \quad \ a_{k,j}+b_{k,j} = 1, \nonumber \\
    & \quad \ a_{k,j}\ge 0, \ b_{k,j} \ge 0.\nonumber
\end{align}

Then, on the basis of the problem in (\ref{lamrepresent01}), we can equivalently rewrite the problem in (\ref{feasibleproject}) as
\begin{align}
    &\min_{\boldsymbol{\delta}^{(i)},\boldsymbol{a},\boldsymbol{b}} \sum_{k\in\mathcal{K}}{\sum_{j\in\mathcal{J}_k}}{\left[ (\delta_{k,j}^{\dag})^2a_{k,j}+(1-\delta_{k,j}^{\dag})^2b_{k,j} \right]} \label{representation02} \\
    & \mathrm{s.t.}\ (\ref{consOfDataSelDelta01}), (\ref{consOfDataSelDelta02}), (\ref{consOfrepresent01}), (\ref{consOfrepresent02}), (\ref{consOfrepresent03}).\nonumber
\end{align}
Note that (\ref{representation02}) is a mixed integer problem and its constraint matrix is totally unimodular. As such, we can relax the binary constraint in (\ref{consOfDataSelDelta01}) to (\ref{consOfDataSelDelta03}) without any loss of optimality, that is, the problems in (\ref{representation02}) and (\ref{lambda-representation}) are equivalent. Given the equivalence between the problems in (\ref{feasibleproject}) and (\ref{representation02}), we can conclude that the problems in (\ref{feasibleproject}) and (\ref{lambda-representation}) are equivalent.
\end{appendices}


\end{document}